\documentclass{article}

\usepackage{hyperref}

\usepackage{alltt}
\usepackage{amsthm}
\usepackage{stmaryrd}
\usepackage{amsmath,amsthm,amssymb}
\usepackage{xparse}
\usepackage{multirow}
\usepackage{amsmath,amsthm,amssymb}
\usepackage{booktabs}
\usepackage{tikz}
\usepackage{emptypage}
\usepackage{esvect}
\usetikzlibrary{arrows,positioning,fit,shapes,calc}
\usepackage{framed}
\usepackage{chessboard}
\storechessboardstyle{5x5}{maxfield=e5}

\usetikzlibrary{calc,angles,positioning,intersections,quotes,decorations.markings}
\usepackage{tkz-euclide}
\usepackage{pgfplots}
\usepackage{graphicx}
\usepackage{subfigure}
\usepackage{float}
\usepackage{multicol}
\pgfplotsset{compat=1.5}

\newcommand{\C}{\mathbb{C}}

\newcommand{\PP}{\mathcal{P}}
\newcommand{\GG}{\mathcal{G}}
\newcommand{\Select}{\mathcal{S}}
\newcommand{\Cross}{\mathcal{C}}
\newcommand{\Mutate}{\mathcal{M}}
\newcommand{\Aim}{\mathcal{A}}
\newcommand{\Divert}{\mathcal{D}}
\newcommand{\Next}{\mathcal{N}ext}
\newcommand{\EvaA}{\mathcal{A}eval}
\newcommand{\EvaB}{\mathcal{B}eval}

\newcommand{\SC}{\mathcal{SC}}

\newcommand{\fd}{\Rightarrow}
\newcommand{\ms}{\mapsto}

\newenvironment{proposition}[2][Proposicion]{\begin{trivlist}
\item[\hskip \labelsep {\bfseries #1}\hskip \labelsep {\bfseries #2.}]}{\end{trivlist}}
\newenvironment{corollary}[2][Corolario]{\begin{trivlist}
\item[\hskip \labelsep {\bfseries #1}\hskip \labelsep {\bfseries #2.}]}{\end{trivlist}}

\pgfplotsset{soldot/.style={color=blue,only marks,mark=*}} \pgfplotsset{holdot/.style={color=blue,fill=white,only marks,mark=*}}

\usepackage{mdframed}
\newdimen\arrowsize
\pgfarrowsdeclare{squarea}{squarea}
{
  \arrowsize=0.4pt
  \advance\arrowsize by.275\pgflinewidth%
  \pgfarrowsleftextend{+-\arrowsize}
  \advance\arrowsize by.5\pgflinewidth
  \pgfarrowsrightextend{+\arrowsize}
}
{
  \arrowsize=0.4pt
  \advance\arrowsize by.275\pgflinewidth%
  \pgfsetdash{}{+0pt}
  \pgfsetroundjoin
  \pgfpathmoveto{\pgfqpoint{1\arrowsize}{4\arrowsize}}
  \pgfpathlineto{\pgfqpoint{-7\arrowsize}{4\arrowsize}}
  \pgfpathlineto{\pgfqpoint{-7\arrowsize}{-4\arrowsize}}
  \pgfpathlineto{\pgfqpoint{1\arrowsize}{-4\arrowsize}}
  \pgfpathclose
  \pgfusepathqfillstroke
}

\pgfarrowsdeclare{open squarea}{open squarea}
{
  \arrowsize=0.4pt
  \advance\arrowsize by.275\pgflinewidth%
  \pgfarrowsleftextend{+-.5\pgflinewidth}
  \advance\arrowsize by7\arrowsize
  \advance\arrowsize by.5\pgflinewidth
  \pgfarrowsrightextend{+\arrowsize}
}
{
  \arrowsize=0.4pt
  \advance\arrowsize by.275\pgflinewidth%
  \pgfsetdash{}{+0pt}
  \pgfsetroundjoin
  \pgfpathmoveto{\pgfqpoint{8\arrowsize}{4\arrowsize}}
  \pgfpathlineto{\pgfqpoint{0\arrowsize}{4\arrowsize}}
  \pgfpathlineto{\pgfqpoint{0\arrowsize}{-4\arrowsize}}
  \pgfpathlineto{\pgfqpoint{8\arrowsize}{-4\arrowsize}}
  \pgfpathclose
  \pgfusepathqstroke
}

\makeatletter
\newdimen\tempa
\newdimen\tempb
\pgfdeclareshape{diamond in circle}{
\inheritsavedanchors[from=diamond]
\inheritsavedanchors[from=circle]
\inheritanchorborder[from=circle]
\inheritanchor[from=circle]{center}
\inheritanchor[from=circle]{radius}
\inheritanchor[from=circle]{north}
\inheritanchor[from=circle]{south}
\inheritanchor[from=circle]{east}
\inheritanchor[from=circle]{west}
\inheritanchor[from=circle]{anchorborder}
  \saveddimen\radius{%
    \pgf@ya=.5\ht\pgfnodeparttextbox%
    \advance\pgf@ya by.5\dp\pgfnodeparttextbox%
    \pgfmathsetlength\pgf@yb{\pgfkeysvalueof{/pgf/inner ysep}}%
    \advance\pgf@ya by\pgf@yb%
    \pgf@xa=.5\wd\pgfnodeparttextbox%
    \pgfmathsetlength\pgf@xb{\pgfkeysvalueof{/pgf/inner xsep}}%
    \advance\pgf@xa by\pgf@xb%
    \pgf@process{\pgfpointnormalised{\pgfqpoint{\pgf@xa}{\pgf@ya}}}%
    \ifdim\pgf@x>\pgf@y%
        \c@pgf@counta=\pgf@x%
        \ifnum\c@pgf@counta=0\relax%
        \else%
          \divide\c@pgf@counta by 255\relax%
          \pgf@xa=16\pgf@xa\relax%
          \divide\pgf@xa by\c@pgf@counta%
          \pgf@xa=16\pgf@xa\relax%
        \fi%
      \else%
        \c@pgf@counta=\pgf@y%
        \ifnum\c@pgf@counta=0\relax%
        \else%
          \divide\c@pgf@counta by 255\relax%
          \pgf@ya=16\pgf@ya\relax%
          \divide\pgf@ya by\c@pgf@counta%
          \pgf@xa=16\pgf@ya\relax%
        \fi%
    \fi%
    \pgf@x=\pgf@xa%
    \pgfmathsetlength{\pgf@xb}{\pgfkeysvalueof{/pgf/minimum width}}%
    \pgfmathsetlength{\pgf@yb}{\pgfkeysvalueof{/pgf/minimum height}}%
    \ifdim\pgf@x<.5\pgf@xb%
        \pgf@x=.5\pgf@xb%
    \fi%
    \ifdim\pgf@x<.5\pgf@yb%
        \pgf@x=.5\pgf@yb%
    \fi%
    \pgfmathsetlength{\pgf@xb}{\pgfkeysvalueof{/pgf/outer xsep}}%
    \pgfmathsetlength{\pgf@yb}{\pgfkeysvalueof{/pgf/outer ysep}}%
    \ifdim\pgf@xb<\pgf@yb%
      \advance\pgf@x by\pgf@yb%
    \else%
      \advance\pgf@x by\pgf@xb%
    \fi%
  }
\backgroundpath{
    \tempa=\radius
    \pgfmathsetlength{\pgf@xb}{\pgfkeysvalueof{/pgf/outer xsep}}%
    \pgfmathsetlength{\pgf@yb}{\pgfkeysvalueof{/pgf/outer ysep}}%
    \ifdim\pgf@xb<\pgf@yb%
      \advance\tempa by-\pgf@yb%
    \else%
      \advance\tempa by-\pgf@xb%
    \fi%
    \pgfpathmoveto{\centerpoint\advance\pgf@x by\radius}%
    \pgfpathlineto{\centerpoint\advance\pgf@y by\radius}%
    \pgfpathlineto{\centerpoint\advance\pgf@x by-\radius}%
    \pgfpathlineto{\centerpoint\advance\pgf@y by-\radius}%
    \pgfpathclose%
  }
\behindbackgroundpath{
    \tempa=\radius%
    \pgfmathsetlength{\pgf@xb}{\pgfkeysvalueof{/pgf/outer xsep}}%
    \pgfmathsetlength{\pgf@yb}{\pgfkeysvalueof{/pgf/outer ysep}}%
    \ifdim\pgf@xb<\pgf@yb%
      \advance\tempa by-\pgf@yb%
    \else%
      \advance\tempa by-\pgf@xb%
    \fi%
    \pgfpathcircle{\centerpoint}{\tempa}%
  }
}
\makeatother

\usepackage{enumitem}
\usepackage{multicol}
\usepackage{proof}
\usepackage{vwcol} 

\usepackage[english]{babel}
\usepackage{amsfonts}
\usepackage{amsmath}
\usepackage{mathtools}
\usepackage{amssymb}
\usepackage{epsfig}
\usepackage{alltt}
\usepackage[latin1]{inputenc}
\usepackage{fixme}

\usepackage{pstricks}
\usepackage{pst-node}

\usepackage{enumerate}
\usepackage[T1]{fontenc}
\usepackage[latin1]{inputenc}
\usepackage{amssymb,amsmath}
\usepackage{color}
\usepackage{hyperref}
\usepackage{graphics}

\newcounter{cdobleimpl}
\def\thecdobleimpl{\ifnum\value{cdobleimpl}=1 $\Longrightarrow$:\ \else $\Longleftarrow$:\ \fi}

\def\squareforqed{\hbox{\rlap{$\sqcap$}$\sqcup$}}
\def\qed{\ifmmode\squareforqed\else{\unskip\nobreak\hfil
\penalty50\hskip1em\null\nobreak\hfil\squareforqed
\parfillskip=0pt\finalhyphendemerits=0\endgraf}\fi}
\let\ab\allowbreak
\newtheorem{theorem}{\bfseries Theorem}
\newtheorem{definition}{\bfseries Definition}
\newtheorem{lemma}{\bfseries Lemma}
\newtheorem{example}{\bfseries Example}

\newcommand{\ismcomment}[1]{}
\newcommand{\bdfn}{\begin{definition} \begin{rm}}
\newcommand{\edfn}{\end{rm}$ $\qed \end{definition}}
\newcommand{\bthm}{\begin{theorem} \begin{rm}}
\newcommand{\ethm}{\end{rm}$ $\qed \end{theorem}}
\newcommand{\bprop}{\begin{proposition} \begin{rm}}
\newcommand{\eprop}{\end{rm}\qed\end{proposition}}
\newcommand{\bcor}{\begin{corollary}\begin{rm}}
\newcommand{\ecor}{\end{rm} \end{corollary}}
\newcommand{\blem}{\begin{lemma} \begin{rm}}
\newcommand{\elem}{\end{rm}\qed\end{lemma}}
\newcommand{\bfact}{\begin{fact} \begin{rm}}
\newcommand{\efact}{\end{rm} \end{fact}}
\newcommand{\bex}{\begin{example} \begin{rm}}
\newcommand{\eex}{\end{rm}$ $\qed  \end{example}}

\newcommand{\bprf}{\begin{IEEEproof}}
\newcommand{\eprf}{\end{IEEEproof}}

\newenvironment{sketch}%
{\nopagebreak[4]\vspace*{0.2em}\noindent{\bf\it Proof Sketch}:\hspace{1ex}}
  {}
\newcommand{\bprfsketch}{\begin{sketch}}
\newcommand{\eprfsketch}{\end{sketch}}

\newcommand{\comen}[1]{}

\newcommand{\letbar}[1]{\mbox{I\kern-0.23em#1}}

\newbox\arriba
\newbox\abajo
\newbox\CaracterInterno
\newbox\CaracterDerecha
\newdimen\anchura
\def\MacrosTranGeneral#1#2#3#4#5#6{%
  \setbox\CaracterInterno=\hbox{\mathsurround=0pt$\mathord#4$}
  \setbox\CaracterDerecha=\hbox{\mathsurround=0pt$\mathord#3$}
  \setbox\arriba=\hbox{$#1#2$}
  \setbox\abajo=\hbox{\mathsurround=0pt%
                      \anchura=\wd\arriba%
                      \advance \anchura by 0.5em%
                      \divide \anchura by \wd\CaracterInterno%
                      \multiply \anchura by \wd\CaracterInterno%
                      \copy\CaracterInterno\kern\SeparacionInternaFlecha
                      \hbox to \anchura{%
                          $\cleaders%
                            \hbox{\kern\SeparacionInternaFlecha\copy\CaracterInterno}
                            \hfill$}%
                      \kern\SeparacionExternaFlecha\copy\CaracterDerecha}
  \mathrel{{\buildrel\vbox{\copy\arriba \kern\SeparacionFlechaArriba} %
    \over{\copy\abajo^{#6}}}_{#5}}
  }
\def\MacrosTranGeneralProp#1#2#3#4#5{\mathchoice%
  {\MacrosTranGeneral{\scriptstyle}{#1}{#2}{#3}{#4}{#5}}
  {\MacrosTranGeneral{\scriptstyle}{#1}{#2}{#3}{#4}{#5}}
  {\MacrosTranGeneral{\scriptscriptstyle}{#1}{#2}{#3}{#4}{#5}}
  {\MacrosTranGeneral{\scriptscriptstyle}{#1}{#2}{#3}{#4}{#5}}}

\def\MacrosNoTran#1{%
  \def\SeparacionInternaFlecha{-0.3em}
  \def\SeparacionExternaFlecha{-0.5em}
  \def\SeparacionFlechaArriba{-3pt}
  \MacrosTranGeneralProp{#1\kern 0.5em}{{\not\rightarrow}}{-}{}{}}

\begin{document}
%


\title{Generalization and Completeness of Stochastic Local Search Algorithms\thanks{This paper was published in Swarm and Evolutionary Computation.
The present version is the author's accepted manuscript. Work partially supported by projects TIN2015-67522-C3-3-R, PID2019-108528RB-C22, and by Comunidad de Madrid as part of the program S2018/TCS-4339 (BLOQUES-CM) co-funded by EIE Funds of the European Union.}}

\author{Daniel~Loscos,
        Narciso~Mart{\'\i}-Oliet,
        and~Ismael~Rodr{\'\i}guez
\thanks{
Dpto. Sistemas Inform\'aticos y Computaci\'on.
Facultad de Inform\'atica.
Universidad Complutense de Madrid. 28040 Madrid, Spain. Narciso~Mart{\'\i}-Oliet and Ismael \hbox{Rodr{\'\i}guez} are also with Instituto de Tecnolog{\'\i}as del Conocimiento.\protect\\
E-mail: {\tt \{dloscos,narciso,isrodrig\}@ucm.es}%
}}





\date{}

\maketitle

\begin{abstract}
We generalize Stochastic Local Search (SLS) heuristics into a unique formal model. This model has two key components: a common structure designed to be as large as possible and a parametric structure intended to be as small as possible. Each heuristic is obtained by instantiating the parametric part in a different way. Particular instances for Genetic Algorithms (GA), Ant Colony Optimization (ACO), and Particle Swarm Optimization (PSO) are presented. Then, we use our model to prove the Turing-completeness of SLS algorithms in general. The proof uses our framework to construct a GA able to simulate any Turing machine. This Turing-completeness implies that determining any non-trivial property concerning the relationship between the inputs and the computed outputs is undecidable for GA and, by extension, for the general set of SLS methods (although not necessarily for each particular method). Similar proofs are more informally presented for PSO and ACO.
\end{abstract}





\section{Introduction}\label{sec:introduction}

Tackling optimization problems is a necessity in virtually all industrial and scientific areas, yet finding the optimal solution is computationally intractable for many of these problems.
\comen{\textcolor{red}{For many NP-hard optimization problems, if P$\neq$NP then no polynomial-time algorithm 
can reach a performance ratio (i.e. worst-case ratio between the cost of returned solutions and the cost of optimal solutions) that is
constant or polynomial with the size of the input, or even always find valid solutions (these are the cases of \hbox{Log-APX-,} \hbox{Exp-APX-,} and NPO-hard problems, respectively~\cite{pas09}).}}%
Thus, we rather focus on looking for reasonable suboptimal solutions for typical problem instances, normally by means of heuristics. Stochastic Local Search (SLS) algorithms~\cite{Hoos2015}, and particularly notable families of them such as Evolutionary and Swarm optimization methods~\cite{evol97,swarm01}, can be good choices to tackle many \comen{\textcolor{red}{(non-NPO-hard)}} optimization problems. In these methods, a set of simple entities interact with each other according to simple rules, changing across iterations and collaboratively constructing and improving solutions to the problem under consideration. For instance, in a Genetic Algorithm~\cite{ga06}, chromosomes (entities) mix with each other and constitute evolving solutions by themselves, whereas in Ant Colony Optimization~\cite{aco99,aco06} some ants (agents) collaboratively draw a candidate solution (a path or set of paths) on a graph while they iteratively traverse it. SLS heuristics in general, and evolutionary and swarm algorithms in particular, have been applied to a wide variety of domains (see e.g. \cite{DCG05,KaA11,QWZ13,pnd09,rrr17}).

During the last years, this field has witnessed an explosion of new methods which, in some cases, have been motivated mainly by the beauty of the natural procedure they try to replicate, rather than by displaying an actually different behavior guaranteeing new interesting mathematical properties during the search~\cite{sor15}.
%
%
Given the huge number of parameters and alternative steps each method admits, making a new algorithm reach good solutions for a known benchmark (or a significant part of it) is often a matter of using a powerful parameter tuner ~\cite{hutter2009paramils,lopez2016irace,hutter2011sequential} to optimize it for the specific benchmark until it achieves its best, sometimes over-fitted, results ---or even just a matter of a long try-and-fail designer time until it works fine (sometimes even by chance, sometimes just because the modifications made it be roughly equivalent to other previously known good algorithm)~\cite{overfitting}. Many newly proposed methods seem to be equivalent to others up to some small changes in some algorithm steps, yielding the feeling that we are speaking about a sort of continuous space of algorithms rather than conceptually different algorithms deserving different names.

Aiming to put some order in this field, some classifications of these algorithms based on their morphological structure, 
natural metaphor, or basic mechanics have been proposed (see e.g.~\cite{dejong,Hoos2015,molina2020comprehensive}), as well as some informally defined generalizations embracing just a few specific algorithms (e.g.~\cite{kang}) and many surveys describing and comparing methods (e.g. \cite{f5,f6}). Perhaps the framework providing more generality is the Generalized Local Search Machine (GLSM)~\cite{hoos2004stochastic}, which models SLS strategies as non-deterministic finite state machines. It illustrates the run-time behaviour and control of SLS algorithms and provides a modelling tool for hybrid and non-cooperative SLS strategies, although it is incomplete: it does not represent termination conditions, neighbourhood relationships, or the definition of the search space or solution set. Additionally, some of the extensions proposed by the authors (namely the ones for cooperative and evolutionary problems) completely alter the framework, instead of slightly modifying some features.

To the best of our knowledge, the literature lacks a fully formal (and general enough) definition of the common and the different parts of SLS methods, such that any method can be the result of taking the framework and instantiating it in some specific way.
Such framework would enable the definition of new and existing algorithms in a common language, easing their classification and direct comparison. By using a common field language, new methods could be developed via a scientific and standardized variation of other known methods, their properties could be more easily investigated, and the results could be seamlessly classified and added into the knowledge base.

In this paper we present a formal language with some fixed structure and some parametric structure such that all SLS methods we are aware of can be the result of instantiating the latter according to the target method.
%
%
Note that, since all (standard or variant) SLS methods are {\it algorithms}, any 
programming language would actually let us implement all of them. Our language, however, will not pursue the generality of programming languages but the opposite: having the {\it least} generality and freedom needed to generalize stochastic local search, in particular by making the common structure be as large and the parametric one be as small as possible. This way, the language aims to make all basic algorithmic elements visible, atomic, and clear for comprehension, manipulation and testing.

%

In order to prove the capability of our general formal model to aid in the discovery of new non-trivial properties, we will use it to prove a disheartening meta-property: the general impossibility of checking any non-trivial property regarding the results of SLS methods. One could argue that this impossibility is an obvious consequence of their randomness: since they involve non-deterministic choices, we cannot predict their result with certainty. However, we will show that their unpredictability lies on deeper reasons far beyond their randomness, as their behavior would remain unpredictable even if they behaved in a deterministic way (or, alternatively, even if we knew in advance the results of all coin flips determining all their random choices during their execution). In particular, this implies that their behavior cannot be predicted {\it in general} by means of statistics, as they would remain unpredictable even in the absence of randomness. 

The key to discover the unpredictability of SLS methods will be proving that they are Turing-complete. Rice's theorem~\cite{rice,cut80} shows that determining if a program written in a Turing-complete language fulfills a non-trivial property regarding only its observable behavior (i.e. the relationship between its inputs and its outputs) is {\it undecidable}.\footnote{By non-trivial we mean fulfilled by neither all programs nor by none. Thus, finding out if programs fulfill properties such as e.g. returning $7$ for input $5$, returning $7$ for some input, computing function $f(x)=2x$, or halting for all inputs is undecidable.}
Our formal model will let us prove that SLS methods are Turing-complete in general, so Rice's theorem will apply to them as well.\footnote{Note that our goal has nothing to do with showing that the {\it result} of an SLS method (i.e. the solution it finds to the problem being solved) can be any program written in a Turing-complete language (the ideal output in Genetic Programming~\cite{koza89,tel94,np08}). Very unrelated to this, we will prove that in general SLS methods {\it are} Turing-complete by themselves, i.e. for any recursive enumerable function, they can compute it.}

One could argue that the Turing-completeness of such methods could be just a trivial consequence of assuming that they are a too general and free concept ---or rather the trivial consequence of representing them with a too general {\it model}. For instance, suppose our general model of SLS methods includes a post-processing step which maps the best solution found by a particular algorithm upon its termination, along with the problem input, into the actual algorithm's solution to be returned to the user. Suppose this post-processing step can be {\it any} algorithm. Then, the Turing-completeness would be the trivial consequence of our freedom to make that step be any algorithm.\footnote{Similarly, by abusing the freedom to define {\it any} fitness function, the fitness function itself could be any computable function.} Quite on the contrary, our proof of Turing-completeness will show that we do not need that much freedom at all ---regardless of whether it is actually allowed by the model. In fact, the Turing-completeness will arise from the intrinsic basic mechanics of these algorithms, even if each involved step is defined in an extremely simple and minimalist way.

In order to emphasize the fact that the generality of the model is not necessary to achieve the Turing-completeness, we will prove the Turing-completeness for a {\it particular} instance of our general model representing a particular SLS method: a Genetic Algorithm (GA). This GA will be such that, given a Turing machine to be simulated and its corresponding input, it iteratively constructs individuals representing longer pieces of the {\it history} (i.e. sequence of traversed configurations) of that Turing machine for that input. The GA will necessarily tend to create such individuals because correctly enlarging that history with subsequent Turing machine steps according to the transitions of the machine will increase their fitness ---although the fitness function will {\it not} embed any Turing machine mechanics and will just check sub-string inclusion. \comen{\textcolor{red}{In order to simplify the construction, we will not directly map the Turing machine and its input into the GA simulating it. On the contrary, our departing point will be a transformation of the Turing machine and its input into an instance of the Modified Post Correspondence Problem (MPCP), as described in~\cite[pp. 401-412]{hopcroft}. The solutions for MPCP instances constructed this way are known to necessarily denote the full history of the Turing machine for that input (which is ultimately the reason of the undecidability of MPCP).}}

Despite the fact that we will show the Turing-completeness of our general model for its GA instantiation in particular, note that the Turing-completeness of a general model is implied by the Turing-completeness of any of its instances, so this will indeed prove the Turing-completeness of SLS {\it in general}. This does not imply that {\it all} particular SLS methods are Turing-complete. Based on our general model, we will also show how our proof, developed in the context of the GA instance, can be easily adapted to other instances representing popular SLS methods, namely Particle Swarm Optimization (PSO)~\cite{pso11} and Ant Colony Optimization (ACO)~\cite{aco99,aco06}.

The contributions of this paper are the following:

\begin{itemize}
\item[(a)] developing the first fully formal model with enough generality to  represent any SLS method we are aware of by instantiating its parametric elements, yet keeping the common part of the model as large as possible;
\item[(b)] using the formal model to prove the inherent unpredictability (in particular, beyond randomness) of 
stochastic local search methods in general, which is due to the Turing-completeness of their mechanics at its most basic level; and
\item[(c)] providing a formal proof of the Turing-completeness of GA, as well as sketching informal proofs for ACO and PSO.
\end{itemize}

%

The rest of the paper is organized as follows. In the next section we describe our general model of SLS heuristics through its operational semantics. Section~\ref{sec:particular} introduces the particular models for GA, PSO, and ACO to exemplify instances of the general form. The proof of the Turing-completeness of GAs (and thus of SLS) is presented in Section~\ref{sec:proof}, and we informally discuss how to prove the Turing-completeness of other instances of 
SLS in Section~\ref{sec:others}. In Section~\ref{sec:conclusions} we present our conclusions and lines of future work.

\section{General Form of stochastic local search}\label{sec:general}

In order to abstract the common structure of SLS strategies, in this section we design a language and a set of operational semantics constituting our  \textit{General Form} of SLS. We claim that Genetic Algorithms, Evolution Strategies, Ant Colony Optimization, Iterated Local Search, Stochastic Gradient Descent, Simulated Annealing, Particle Swarm Optimization and every other stochastic local search algorithm we are aware of are instances of this general operational semantics.

The decisions on how to formalize the semantics will be heavily inspired by the operational semantics of~\textbf{While} by Nielson \& Nielson \cite[pp. 12-14 and 32-36]{nielson}.
More specific semantics will be provided later for different specific algorithms, but we believe the following semantics are at the lowest abstraction level able to generalize the whole of SLS computation. In the following, we assume that a given algorithm \textit{A} receives a specific instance \textit{p} of a problem and explores different solutions. The best solution found by the computation is stored in a variable named \textit{Best}.

\subsection{State and Syntactic Categories}

The \textbf{State} of the computation is a function from variables to values. The relevant variables of the computation constitute the following tuple:\footnote{Other auxiliary variables such as array indexes and loop counters are not reflected here, although they would be necessary for the implementation of SLS algorithms.}
\begin{center}
$(Prob, Sols, Best, FPW, Extra) \in  \textbf{Problem} \times \textbf{Solution[]} \times \textbf{Solution} \times \textbf{F/P/W} \times \textbf{Extra}$
\end{center}

Since we are aiming to generalize the SLS computational model, the categories that define the tuple will be abstract enough to fit any SLS strategy. However, we shall define what each category represents and give a glimpse of its structure in different  algorithms.

\textbf{State} is not a constant function, as the value associated with each variable may change during the computation. The letter $s$ will represent an instance of \textbf{State}. We shall write $s[A \ms b]$ to represent a new state where the value associated to $A$ is $b$ and every other variable has the same value as in the state~$s$.

Next we explain the syntactic categories necessary for SLS computation:

\textbf{Problem} must be able to encapsulate the optimization problem that our algorithm must solve. Not only the abstract problem (e.g. nonlinear optimization, TSP\footnote{Travelling Salesperson Problem: finding the minimally valued route in a positive-valued graph that goes through every single node exactly once.}, etc.) but also the concrete instance and parameters of the problem (e.g. the function to optimize, the graph with edge costs 
for TSP, etc.). When \textit{Prob} $\in$ \textbf{Problem} is initialized at the beginning of the computation, it will determine how the functions required for the computation of the problem work.

\textbf{Solution} is the category that represents possible solutions of the problem within the computation. For instance, in a GA \textbf{Solution} would represent the codification of the chromosomes; and in an ACO algorithm, \textbf{Solution} would be the data type to store the subgraphs that represent the paths followed by ants. \textbf{Solution[]} represents a set of instances of \textbf{Solution}. Thus, \textit{Best} $\in$ \textbf{Solution} will be the variable that represents the best solution found by the computation and \textit{Sols} $\in$ \textbf{Solution[]} the set of solutions that are being considered in the current iteration of the algorithm.

The most general category is \textbf{F/P/W}, which stands for \textit{Fitness / Pheromones / Weights}. \textbf{F/P/W} must be general enough to represent every variable necessary to guide the search in an SLS algorithm; e.g. in a swarm-based heuristic, \textit{FPW }$ \in $\textbf{ F/P/W} would  consist of the Weights of each individual, its linear moment, etc.; for a GA, \textit{FPW} would store the Fitness of every individual, and in an ACO algorithm the Pheromones of each path.

Lastly we introduce the \textbf{Extra} category. It will be used to wrap the parameters required to define the auxiliary functions needed for the computation, as well as any other variable needed by the specific algorithm that is being run. For example, the crossover or mutation rate for a GA would belong to this category. 
\textbf{Extra} will be key to the implementation of complex heuristics, as it is the default way to carry lateral effects of computation steps. In hybrid methods, \textbf{Extra} will carry all the necessary data to transition between heuristics. The functions would read this information and operate accordingly.

Additionally, let \textbf{T} \cite[p.14]{nielson} consist of the truth values \textbf{tt} (for true) and \textbf{ff} (for false), let \textbf{Pexp} be the syntactic category used to input the problem into the computation (an instance of \textbf{Pexp} will be translated into its corresponding \textbf{Problem} instance to be processed), let \textbf{Stm} be the set of statements (semantic blocks) that build the SLS computation model, and \textbf{Algorithm} be the set of SLS algorithms.

We shall also define the meta-variables that will be used to range over constructs of our syntactic categories:

\textit{p} will range over input problem expressions, \textbf{Pexp}.

\textit{A} will range over the set of SLS algorithms, \textbf{Algorithm}. Note that we are considering the abstract algorithms, which are independent of the instance being run or the problem at hand.

Lastly, \textit{S} will range over statements, \textbf{Stm}. Also $S'$, $S_1$ and $S_2$ will stand for statements.

We assume that the structure for \textbf{Pexp} constructs is given elsewhere as it is not relevant for the computation. However, the structure for \textbf{Stm} constructs is indeed relevant and hereby presented:

$S := S_1;S_2 \text{ } | \texttt{ setProb( }p \texttt{ ) } | \texttt{ generate } | \ab \texttt{ nextGen } | \texttt{ evaluate } | \texttt{ stop } | \\\hspace*{4em} \texttt{ compute( }p \texttt{ )}$

We use 
\texttt{compute( }p \!\!\texttt{ )}
to start the execution of our given algorithm \textit{A} by running \texttt{setProb( }p\!\! \texttt{ )}, which translates the problem from \textbf{Pexp} to \textbf{Problem} and then calls \texttt{generate} to create the original set of candidate solutions, \texttt{evaluate} to rate them, and \texttt{stop} to decide whether the algorithm has finished or if the process must be iterated by calling \texttt{nextGen}, which generates a new set of solutions with the acquired SLS knowledge.

The meaning of the statements is further detailed by the following functions and rules.

\subsection{Auxiliary Functions}\label{sec:auxiliary}

Before we begin to introduce the auxiliary functions that will help us define how the state varies during the computation, let us introduce the following notation: Let \textbf{Space} be any space relevant to the computation where variables or syntactic constructs may range (e.g. \textbf{F/P/W}) and let \textbf{Space} not be \textbf{Extra}; then, we define \textbf{Space*} as $\textbf{Space} \times \textbf{Extra}$.

Since any step of the computation and any change of the state may involve a change in \textbf{Extra} for at least some SLS algorithm, we note 
that every one of the following functions can also alter the value of \textbf{Extra}. Next we present the functions:

\begin{itemize}
\item $\C\llbracket A \rrbracket: \textbf{Algorithm} \longrightarrow \textbf{ Extra }$.
This is the function that takes an instance \textit{A} of \textbf{Algorithm} and starts the computation by initializing every parameter and variable needed by the heuristic.
\item $\PP\llbracket p \rrbracket_s: \textbf{Pexp } \times \textbf{ State} \longrightarrow \textbf{ Problem* }$.
This is the semantic function for \textbf{Pexp} that translates the problem from the input form to the computing form that can be stored as a variable value. The previous state of the computation is used to adapt the problem to the algorithm that is being run.
\item $\GG\llbracket \rrbracket_s: \textbf{State } \longrightarrow \textbf{ Solution[]* }$.
This function uses the value of \textit{Prob} given by the state $s$ to generate a new random value for \textit{Sols} concordant to the problem stored.
\item $\Next\llbracket \rrbracket_s: \textbf{State } \longrightarrow \textbf{ Solution[]* }$.
It uses the values of \textit{Prob},  \textit{Sols}, and  \textit{FPW} given by the state $s$ to stochastically compute the new value for \textit{Sols}, using the information obtained by the previous iterations of the algorithm, and in such a way that the solutions are concordant to the problem.
\item $\EvaA\llbracket \rrbracket_s: \textbf{State } \longrightarrow \textbf{F/P/W* }$.
It uses the values of \textit{Prob},  \textit{Sols}, and  \textit{FPW} given by the state $s$ to compute the new value for \textit{FPW}, determined by the problem stored and the information obtained by the previous iterations of the algorithm.
\item $\EvaB\llbracket \rrbracket_s: \textbf{State } \longrightarrow \textbf{ Solution* }$.
It uses the values of \textit{Prob},  \textit{Sols}, \textit{Best}, and \textit{FPW} given by the state $s$ to compute the new value for \textit{Best}, which is the best solution found so far.
\item $\SC \llbracket \rrbracket_s: \textbf{State } \longrightarrow \textbf{ T }$.
This is the stop criterium function. It analyses the whole state and returns \textbf{tt} if the stop criteria has been met and \textbf{ff} if it has not.
\end{itemize}

Since this is an abstraction of the SLS computation model, different SLS algorithms will implement these functions differently. Just to give an example, let us consider two 
GAs: the first one returns the value of the best individual of the last generation, and the second one the best individual amongst all generations. The former has a $\EvaB\llbracket \rrbracket_s$ function that ignores \textit{Best}, whereas the latter does consider it in every generation. Similarly, the parameters of the algorithm directly affect most of these functions.

Readers should also be aware that many of these functions, specially functions $\EvaA\llbracket \rrbracket_s, \EvaB\llbracket \rrbracket_s$ and $\SC\llbracket \rrbracket_s$, could be overloaded with \textit{daemon actions} such as local search phases, automatic parameter modifications, and other mechanisms to improve solutions. For the sake of generality and notation simplicity, no terms or functions are given to {\it specifically} denote daemon actions in our design, as they can be easily integrated within the other elements of the language.

\subsection{Rules}
Finally, we introduce the rules of our SLS computation model in its General Form.
The execution begins at an initial state \textit{s} with the $[\text{compute}]$ rule by inputting the problem \textit{p} as a parameter: $\langle \texttt{compute( } p \texttt{ )}, s \rangle$.

The values of the initial state \textit{s} are not relevant, as they will be changed by the input problem and no previous information will affect the computation in any way.

We introduce a special notation in the same way as we did for defining the functions. Let \textbf{Var} be any variable of \textbf{State} other than \textbf{Extra}. We define \textbf{Var*} as $\textbf{Variable} \times \textbf{Extra}$.

\begin{figure*}[t]
\hrule

\begin{center}
\scriptsize{}

\begin{vwcol}[widths={0.2,0.8}, sep=.8cm, justify=flush,rule=0pt,indent=1em]

$[ \text{comp}^1 ]$\\\\

$ \infer{\langle S_1; S_2, s \rangle \fd \langle S_1'; S_2, s' \rangle}{\langle S_1, s \rangle \fd \langle S_1', s' \rangle}$

\end{vwcol}
\begin{vwcol}[widths={0.2,0.8}, sep=.8cm, justify=flush,rule=0pt,indent=1em]

$[ \text{comp}^2 ]$\\\\

$ \infer{\langle S_1; S_2, s \rangle \fd \langle S_2, s' \rangle}{\langle S_1, s \rangle \fd  s'}$

\end{vwcol}
\begin{vwcol}[widths={0.2,0.8}, sep=.8cm, justify=flush,rule=0pt,indent=1em]

$[ \text{set problem} ]$\\\\

$ \langle \texttt{setProb( } p \texttt{ )}, s \rangle \fd s[Prob^* \ms \PP\llbracket p \rrbracket_s]$\\

\end{vwcol}
\begin{vwcol}[widths={0.2,0.8}, sep=.8cm, justify=flush,rule=0pt,indent=1em]

$[ \text{generate} ]$\\\\

$ \langle \texttt{generate}, s \rangle \fd s[Sols^* \ms \GG\llbracket \rrbracket_s]$\\

\end{vwcol}
\begin{vwcol}[widths={0.2,0.8}, sep=.8cm, justify=flush,rule=0pt,indent=1em]

$[ \text{next generation} ]$\\\\

$ \langle \texttt{nextGen}, s \rangle \fd s[Sols^* \ms \Next\llbracket \rrbracket_s]$\\

\end{vwcol}
\begin{vwcol}[widths={0.2,0.8}, sep=.8cm, justify=flush,rule=0pt,indent=1em]

$[ \text{evaluate} ]$\\\\

$\langle \texttt{evaluate}, s \rangle \fd s[FPW^* \ms \EvaA\llbracket \rrbracket_s, Best^* \ms \EvaB\llbracket \rrbracket_s]$

\end{vwcol}
\begin{vwcol}[widths={0.2,0.8}, sep=.8cm, justify=flush,rule=0pt,indent=1em]

$[ \text{stop}^\text{tt} ]$\\

$\langle \texttt{stop}, s \rangle \fd s
\qquad\qquad\qquad\qquad\qquad\qquad
\text{ if } \SC \llbracket \rrbracket_s = \textbf{tt}$

\end{vwcol}
\begin{vwcol}[widths={0.2,0.8}, sep=.8cm, justify=flush,rule=0pt,indent=1em]

$[ \text{stop}^\text{ff} ]$\\

$\langle \texttt{stop}, s \rangle \fd \langle \texttt{nextGen}; \texttt{evaluate}; \texttt{stop}, s \rangle  \quad \text{ if } \SC \llbracket \rrbracket_s = \textbf{ff}$

\end{vwcol}
\begin{vwcol}[widths={0.2,0.8}, sep=.8cm, justify=flush,rule=0pt,indent=1em]

$[ \text{compute} ]$\\\\

$ \infer{\langle \texttt{compute( } p \texttt{ )}, s \rangle \fd \langle \texttt{generate}; \texttt{evaluate}; \texttt{stop}, s' \rangle}{\langle \texttt{setProb( } p \texttt{ )}, s[Extra \ms \C\llbracket A \rrbracket ] \rangle \fd  s' }$

\end{vwcol}
\end{center}
\hrule\caption{Operational semantics for the General Form.}\label{fig:general}
\end{figure*}

As usual in operational semantics, the meaning of the statements is specified by a transition system with two kinds of configurations:

\begin{itemize}
\item $\langle S, s \rangle$ represents that the statement $S$ is to be executed from the state $s$.

\item $s$ represents a terminal state. These final states often come in the $s[A \ms b]$ notation.
\end{itemize}

A valid transition from configuration $\alpha $ to configuration $\beta $ is represented by $\alpha \fd \beta $. The transition rule $\begin{array}{l} \infer{\alpha \fd \beta}{\gamma \fd \delta }\end{array}$ indicates that we can only transition from configuration $\alpha $ to configuration $\beta $ if a valid transition from $\gamma $ to $\delta$ can be made.

The rules for the General Form are defined in Figure~\ref{fig:general} and have the following purposes: $[ \text{comp}]$ rules allow to concatenate instructions carrying on the state modifications;  $[ \text{set problem} ]$ loads the instance to run onto the state of the computation; $[ \text{generate} ]$, $[ \text{next generation} ]$, and $[ \text{evaluate} ]$ run their corresponding statements to generate the initial solutions, the following generation, and evaluating the fitness of their generation, respectively; and the $[ \text{stop} ]$ rules either finish the computation and return the final state or trigger the next iteration of the process, depending on the evaluation of the stop function at the current state. Finally $[ \text{compute} ]$ is the first rule we apply to start the computation, and all the others are applied as we unravel~it.

\section{GAs, PSO \& ACO as instances of the General Form}\label{sec:particular}

Instances of the General Form for particular algorithms (GA, ACO, PSO) are introduced in this section.

\subsection{Genetic Algorithms}
Now that the general model has been presented, more detailed semantics, based on it, will be provided for genetic algorithms. The goal is now to showcase the structure of the computation of genetic algorithms while keeping the semantics general enough to be applied to particular instances of the genetic algorithms (i.e. different variants of GA). The representation of the chromosomes, selection method, mutation rate, 
stop criteria, and other particularities of the genetic algorithm instance should not be relevant to the semantics. This semantics is general enough to cover degenerated instances of the genetic algorithm such as e.g. evolutionary strategies, where the selection and crossover steps are trivial.

The statements and functions added to this semantics are meant to show the importance of sequentially modifying a set of individuals to guide the search. The memory of genetic algorithms resides on its population, as fitness values of a generation are no longer relevant after the creation of the next generation.


The \textbf{State} is defined as in the General Form. However, some syntactic categories can be further specified.
In this case, \textbf{Solution} will represent the codification of a single chromosome and \textbf{Solution[]} a set of instances of \textbf{Solution}. Thus, \textit{Best} $\in$ \textbf{Solution} will be the fittest chromosome found by the computation and \textit{Sols} $\in$ \textbf{Solution[]} the current generation together with every other solution needed to compute crossovers and mutations. Also, \textit{FPW} $ \in $ \textbf{F/P/W} now consists of the fitness of every individual (which must be recomputed for every generation).

We will use the same meta-variables of the General Form, the only difference being the structure for \textbf{Stm} constructs:

$S := S_1;S_2 \text{ } | \texttt{ setProb( }p \texttt{ ) } | \texttt{ generate } | \ab \texttt{ select } |\texttt{ cross } | \texttt{ mutate } | \\\hspace*{4em}\texttt{ nextGen } | \ab \texttt{ evaluate } | \texttt{ stop } | \texttt{ compute( }p \texttt{ )}$

Three new statements (\texttt{select}, \texttt{cross}, and \texttt{mutate}) are considered to structure the process that was previously abstracted by \texttt{nextGen} in a single step. This showcases how the evolution is performed in GAs: fitness-biased selection of individuals to cross over, generation of new individuals, and mutation of the population. The meaning of the statements is further specified by the following functions and rules.


\begin{itemize}
\item $\Select\llbracket \rrbracket_s: \textbf{State } \longrightarrow \textbf{ Solution[]* }$.
This uses the values of \textit{Prob},  \textit{Sols}, and \textit{FPW} given by the state $s$ to compute the new value of \textit{Sols} consisting of the individuals of the previous generation selected for the crossover stage.
\item $\Cross\llbracket \rrbracket_s: \textbf{State } \longrightarrow \textbf{ Solution[]* }$.
It uses the values of \textit{Prob},  \textit{Sols}, and \textit{FPW} given by the state $s$ to compute the new value of \textit{Sols}, which is determined by a combination of  the result of the crossover operation applied to the selected individuals and some of the original individuals.
\item $\Mutate\llbracket \rrbracket_s: \textbf{State } \longrightarrow \textbf{ Solution[]* }$.
It uses the values of \textit{Prob},  \textit{Sols}, and  \textit{FPW} given by the state $s$ to compute the new value of \textit{Sols} consisting of the individuals already present in \textit{Sols} after modifying (mutating) some of them.
\end{itemize}

Since this is an abstraction of the genetic algorithm computation model, different genetic algorithms will implement these functions differently. For instance, consider a GA with elitism where, in each generation, some of the individuals will be selected as the elite by $\Select\llbracket \rrbracket_s$; the $\Cross\llbracket \rrbracket_s$ operation will force them into the next value of \textit{Sols} and $\Mutate\llbracket \rrbracket_s$ will not modify them. On the contrary, in an elite-less GA every individual will be subject to modifications by $\Cross\llbracket \rrbracket_s$ or $\Mutate\llbracket \rrbracket_s$, and may not be selected by $\Select\llbracket \rrbracket_s$. The definitions of the other functions used in the following rules remain unmodified with respect to the General Form.

To complete the semantics, we take the rules from the General Form and modify them by redefining one rule ($[ \text{next generation} ]$) and adding three new ones: $[ \text{selection} ]$, $[ \text{crossover} ]$, and $[ \text{mutation} ]$ are introduced to represent the corresponding namesake stages where a GA performs its population transformations in each generation. The changes over the operational semantics of the General Form are presented in Figure~\ref{fig:ga}.

\begin{figure}
\hrule
\scriptsize{}

\begin{vwcol}[widths={0.2,0.8}, sep=.8cm, justify=flush,rule=0pt,indent=1em]

$[ \text{selection} ]$\\\\

$ \langle \texttt{select}, s \rangle \fd s[Sols^* \ms \Select\llbracket \rrbracket_s]$\\

\end{vwcol}
\begin{vwcol}[widths={0.2,0.8}, sep=.8cm, justify=flush,rule=0pt,indent=1em]

$[ \text{crossover} ]$\\\\

$ \langle \texttt{cross}, s \rangle \fd s[Sols^* \ms \Cross\llbracket \rrbracket_s]$\\

\end{vwcol}
\begin{vwcol}[widths={0.2,0.8}, sep=.8cm, justify=flush,rule=0pt,indent=1em]

$[ \text{mutation} ]$\\\\

$ \langle \texttt{mutate}, s \rangle \fd s[Sols^* \ms \Mutate\llbracket \rrbracket_s]$\\

\end{vwcol}
\begin{vwcol}[widths={0.2,0.8}, sep=.8cm, justify=flush,rule=0pt,indent=1em]

$[ \text{next generation} ]$\\\\

\;\;\;\;\hbox{$ \langle \texttt{nextGen}, s \rangle \fd \langle \texttt{select}; \texttt{cross}; \texttt{mutate}, s' \rangle$}
\end{vwcol}

\vspace*{1em}

\hrule\caption{Instantiation of the operational semantics for Genetic Algorithms.}\label{fig:ga}
\end{figure} 
\subsection{Ant Colony Optimization}

In this section we particularize the operational semantics from the General Form to represent the Ant Colony Optimization algorithm. ACO algorithms generate a completely new population on each generation. However, the previous iterations influence how the new generation is created. We may, therefore, establish that the memory of ACO resides on the pheromones rather than on the individuals.


For this model, the \textbf{State} is also defined as it was in the General Form, although some of the syntactic categories are further specified as follows.

\textbf{Problem} is basically as described in the General Form. The only variation is that now we are dealing with some form of optimal routing search in a graph, so it will at least represent the graph of the problem as well as other relevant data for the computation.

\textbf{Solution} is the category that represents possible solutions of the considered route finding problem, so it is a data type to store ordered subgraphs. These subgraphs are formed from the paths taken by ants. Note that sometimes the best solution may be the composition of paths traversed by several different ants. \textbf{Solution[]} represents a set of instances of \textbf{Solution}. Thus, \textit{Best} $\in$ \textbf{Solution} will be the most efficient subgraph found by the computation and \textit{Sols} $\in$ \textbf{Solution[]} the set of subgraphs generated and simulated by ants in the current iteration of the algorithm.

The core of \textbf{F/P/W} will be the representation of the pheromones dropped by the ants of previous iterations. Thus, \textit{FPW} $\in$ \textbf{F/P/W} will store optimality values for more than one iteration. Additionally, it will represent any performance-based parameters needed for the computation of auxiliary functions. 
Of course, the pheromone values do not need to be scalar, and vectors are a natural way to implement multi-pheromone ACO heuristics ~\cite{ska15}.

We will use the same meta-variables of the General Form, the only difference being the structure for \textbf{Stm} constructs, now given by:

$S := S_1;S_2 \text{ } | \texttt{ setProb( }p \texttt{ ) } | \texttt{ generate } | \ab \texttt{ nextGen } | \texttt{ simulate } | \\\hspace*{4em}\texttt{ evaluate } | \texttt{ stop } | \ab \texttt{ compute( }p \texttt{ )}$

In this case, the only new statement is \texttt{simulate}. This decision is made to highlight the relevance of the fitness evaluation for these methods. Although the generation of new individuals is pheromone-biased and relatively simple, it is in the fitness evaluation of the chosen paths (simulation) where pheromones are dropped for the following generation. To make use of this statement, we introduce the $[ \text{simulate} ]$ rule and modify $[ \text{evaluate} ]$ to include the simulation step.

The list of functions remains unchanged from the General Form, but their specifications are worth noting:


\begin{itemize}

\item $\Next\llbracket \rrbracket_s: \textbf{State } \longrightarrow \textbf{ Solution[]* }$. This uses the values of \textit{Prob} and  \textit{FPW} given by the state $s$ to compute the new value of \textit{Sols} determined by the stored problem and the information obtained by the previous iterations of the algorithm (except for the first iteration). Note that \textit{Sols} is no longer relevant for this function.
\item $\EvaA\llbracket \rrbracket_s: \textbf{State } \longrightarrow \textbf{F/P/W* }$. It is as seen in the General Form, although in this particular case this function carries out two different tasks: computing the fitness of every ant in \textit{Sols} and storing the pheromones dropped by it.
\end{itemize}

Again, different ant colony algorithms may implement these functions differently; e.g.: $\EvaB\llbracket \rrbracket_s$ could consider solutions not present in \textit{Sols}, such as the densest pheromone path, as potential best solutions. Finally, the rules for Ant Colony Optimization are the same as the ones of the General Form with the modifications mentioned to include  \texttt{simulate}, as shown in Figure~\ref{fig:aco}.

\begin{figure}
\hrule
\vspace*{1em}
\scriptsize{}

\begin{vwcol}[widths={0.2,0.8}, sep=.8cm, justify=flush,rule=0pt,indent=1em]

$[ \text{evaluate} ]$\\

$ \infer{\langle \texttt{evaluate}, s \rangle \fd s'[Best^* \ms \EvaB\llbracket \rrbracket_s]}{\langle \texttt{simulate}, s \rangle \fd  s' }$

\end{vwcol}
\begin{vwcol}[widths={0.2,0.8}, sep=.8cm, justify=flush,rule=0pt,indent=1em]

$[ \text{simulate} ]$\\\\

$\langle \texttt{simulate}, s \rangle \fd s[FPW^* \ms \EvaA\llbracket \rrbracket_s]$

\end{vwcol}
\hrule\caption{Instantiation of the operational semantics for Ant Colony Optimization}\label{fig:aco}
\end{figure} 
\subsection{Particle Swarm Optimization}

The last particularization of the operational semantics for SLS computation we present in this paper is the operational semantics for Particle Swarm Optimization algorithms.

In PSO individuals of the solution population move around the search space trying to find the optimal solution. This process is similar to that of GAs, since we can see each movement as a new generation of solutions generated after the previous population by following a set of rules. However, the way the new positions of the particles are computed is structurally different from that of GAs.
Again, the memory of PSO lies on the individuals and not so much on the weights that have to be recalculated on each generation.
Thus, this is an interesting example to illustrate how the same philosophy of GAs can be implemented in a different computational model and still have its operational semantics included in our General Form of SLS computation.


The \textbf{State} is defined as it was in the General Form, although some syntactic categories are further specified as follows.

 In this case, \textbf{Solution} represents the position, direction, and speed of a particle in the search space, and \textbf{Solution[]} represents a set of \textbf{Solution} particles. Thus, \textit{Best} $\in$ \textbf{Solution} will consist in the optimal point found by the computation so far (ignoring its momentum) and \textit{Sols} $\in$ \textbf{Solution[]} will give the current position and momentum of each particle of the swarm.

The \textbf{F/P/W} category represents every variable necessary to direct the search in the SLS scheme. \textit{FPW} $ \in $ \textbf{F/P/W} now consists of the fitness-based weights of every particle (which must be recomputed in every generation).

We will use the same meta-variables of the General Form, the only difference being the structure for \textbf{Stm} constructs, now given by:

$S := S_1;S_2 \text{ } | \texttt{ setProb( }p \texttt{ ) } | \texttt{ generate } | \ab \texttt{ divert } |\texttt{ aim } | \texttt{ move } | \\\hspace*{4em}\texttt{ nextGen } | \texttt{ evaluate } \ab | \ab\texttt{ stop } | \texttt{ compute( }p \texttt{ )}$

\textbf{Stm} is changed in a similar fashion as it was for GA: \texttt{divert}, \texttt{aim}, and \texttt{move} are introduced to detail the way in which \texttt{nextGen} works. First, \texttt{divert} introduces a random influence in the future movement, then \texttt{aim} targets the movement towards the objective, and finally \texttt{move} combines those two forces to set the new position of the particle.

The order of \texttt{divert} and \texttt{aim} statements is actually interchangeable for the computation, but for the functions and rules that follow, we will assume that \texttt{divert} precedes \texttt{aim} in every iteration. To make use of these statements, we will add their namesake rules. Similarly as for GA, additional functions are required:


\begin{itemize}
\item $\Divert\llbracket \rrbracket_s: \textbf{State } \longrightarrow \textbf{ Solution[]* }$. This function uses the values of \textit{Prob} given by the state $s$ to compute new random momentum values for \textit{Sols} while not changing any positions. These momentum changes are called diversions.
\item $\Aim\llbracket \rrbracket_s: \textbf{State } \longrightarrow \textbf{ Solution[]* }$. It uses the values of \textit{Prob},  \textit{Sols}, and \textit{FPW} given by the state $s$ to compute attractions between particles of \textit{Sols}, then merges this attraction with the momentum values obtained by $\Divert\llbracket \rrbracket_s$ and sets a new momentum value for each particle in \textit{Sols}.
\item $\Mutate\llbracket \rrbracket_s: \textbf{State } \longrightarrow \textbf{ Solution[]* }$. It uses the values of \textit{Sols} given by the state $s$ to compute the new positions of every particle in \textit{Sols} based on its previous position and momentum (direction and speed).
\item $\EvaA\llbracket \rrbracket_s: \textbf{State } \longrightarrow \textbf{F/P/W* }$. This is as seen in the General Form although, in this particular case, the function computes and updates the fitness values of the individuals based on their position inside the search space. The fitness of every particle determines its weight for the attraction stage.
\end{itemize}

Once more, different particle swarm algorithms may implement these functions differently. 
For example, different neighbourhood topologies ~\cite{lwy16} would yield different $\Divert\llbracket \rrbracket_s$ and $\Aim\llbracket \rrbracket_s$ functions.
Similarly as in the previous instances, the list of rules for PSO adds new ones to showcase the stages of the computation in each iteration, as well as a modification in $[ \text{next generation} ]$\ to introduce them, as it is depicted in Figure~\ref{fig:pso}.

\begin{figure}
\hrule
\scriptsize{}

\begin{vwcol}[widths={0.2,0.8}, sep=.8cm, justify=flush,rule=0pt,indent=1em]

$[ \text{divert} ]$\\\\

$ \langle \texttt{divert}, s \rangle \fd s[Sols^* \ms \Divert\llbracket \rrbracket_s]$\\

\end{vwcol}
\begin{vwcol}[widths={0.2,0.8}, sep=.8cm, justify=flush,rule=0pt,indent=1em]

$[ \text{aim} ]$\\\\

$ \langle \texttt{aim}, s \rangle \fd s[Sols^* \ms \Aim\llbracket \rrbracket_s]$\\

\end{vwcol}
\begin{vwcol}[widths={0.2,0.8}, sep=.8cm, justify=flush,rule=0pt,indent=1em]

$[ \text{move} ]$\\\\

$ \langle \texttt{move}, s \rangle \fd s[Sols^* \ms \Mutate\llbracket \rrbracket_s]$\\

\end{vwcol}
\begin{vwcol}[widths={0.2,0.8}, sep=.8cm, justify=flush,rule=0pt,indent=1em]

$[ \text{next generation} ]$\\\\

$ \langle \texttt{nextGen}, s \rangle \fd \langle \texttt{divert}; \texttt{aim}; \texttt{move}, s' \rangle$

\end{vwcol}
\hrule\caption{Instantiation of the operational semantics for Particle Swarm Optimization}\label{fig:pso}
\end{figure} 

The previous three examples (GA, ACO, PSO) are illustrative of the power of the General Form semantics to model the semantic behavior of different algorithms of stochastic local search. We will use them to prove a significant semantic property: the Turing-completeness of SLS strategies in general. In the next section, we will prove this property through one of the instances of our SLS model.

Note that, given that Turing-completeness,  {\it any} algorithm of any kind (not just SLS) can be emulated by our SLS model ---if properly codified as input of the SLS instance proved to be Turing-complete. We trivially infer that, in particular, any {\it SLS algorithm} can be codified into and emulated by our SLS model. Yet the purpose of our SLS model is not just to enable the representation of any SLS method, but also doing so through a natural use of the model parameters. The model instances defined in the previous sections for GA, ACO, and PSO illustrate this capability, and  similarly natural instances can be created for variants of these methods and others.

We already mentioned how the framework could deal with PSO algorithms with different neighbourhood relationships, as well as with multi-pheronome ACO algorithms. Clearly, algorithms originally designed as a modification of others are expected to be defined in a similar way (e.g. River Formation Dynamics (RFD)~\cite{rrrUC07} could be defined in a very similar way as ACO).

Many popular algorithm variations are easy to define as well. For instance, for multi-population algorithms~\cite{can99}, a population identifier can be added as a parameter for individuals and be handled by the different functions. For multi-modal or multi-objective optimization~\cite{dmq11,ma04,laumanns2004running}, $Best$ may easily become a vector, and $\EvaB\llbracket \rrbracket_s$ can manage the identification of different objectives. In co-evolutionary algorithms~\cite{jws13}, we could borrow the previous techniques to handle multi-population, and $Extra$ could include the structures needed to manage the global fitness of the model. Similar techniques can also cover the instantiation of decomposition-based strategies~\cite{zl07}. Some trajectory-based heuristics such as Simulated Annealing~\cite{kirkpatrick1983osa} or Iterated Local Search require a single individual population. Our General Form also accommodates techniques to handle constrained optimization problems~\cite{kb07}, in particular by including penalization or blacklisting parameters into $\EvaA\llbracket \rrbracket_s$ (independently of the heuristic being instanced).

Of course, these techniques for the mentioned heuristics are not the only ways to instantiate our framework  ---they are just an illustration of its versatility. As an final example, consider the SEMO, FEMO and GEMO multi-objective algorithms presented in~\cite{laumanns2004running}: All of them keep a population of Pareto optimal individuals, therefore making $Best = Sols$ at the end of each iteration. Here we see how having $Best$ be a vector easily accommodates for multi-objective optimization. We can also divide the ``next generation'' stage of these methods in two steps: selection and mutation. Since the three algorithms only differ in which individuals are mutated, their instantiation would be the same except for a slightly different $\Mutate\llbracket \rrbracket_s$ function. This way our model captures their common general structure while adapting for their particularities.

\section{Turing-completeness of stochastic local search heuristics}\label{sec:proof}

In this section we prove the Turing-completeness of SLS computation, which means that any program that can run on a Turing Machine (TM) can also be run by launching an SLS strategy. We achieve this by proving the Turing-completeness of an instance of these strategies: genetic algorithms. As far as we know, this property has not been proved before and has serious implications for the investigation of semantic properties.
Note that the convergence of some SLS methods has been studied in the literature under specific conditions and configuration settings (see e.g.~\cite{rudolph1998finite,mitra1986convergence,johnson2002convergence,tre03,cq07,jly07}). Due to the Turing-completeness of SLS methods and Rice's theorem, we conclude that very strong and exceptional hypotheses will be required, in general, to achieve this kind of semantic knowledge ---in the same way as proving that a program produces its expected outputs is undecidable in general, but it can be proved for some subsets of programs if we heavily constrain their form in some way (e.g. programs with restricted loop structures or no loops at all).

Introducing strong constraints can let us reason about desirable semantic properties in some SLS settings; e.g.  polynomial-time approximation to maximizing submodular functions ~\cite{qian2019maximizing,friedrich2015maximizing} or convergence of SA to the global optimum under strong ergodicity \cite{mitra1986convergence}. However, restrictions on the expressivity of the method (when viewed as a computation model) will always be come with them. Turing-completeness also disables {\it in general} the use Statistics to predict the behavior of SLS methods. Even if we defined a GA that simulated a universal Turing machine, run it for many instances and calculated its average outputs or execution times; this would not tell us anything about what to expect, in general, if the same GA was run for new untested instances. It would be equivalent to trying to predict the behavior of a C++ program for all inputs just by observing what it does for some and calculating the average of its outputs or execution times. Statistics are, however, very useful indeed for the analysis of SLS methods when additional constraints do limit their scope.

Note that proving the  Turing-completeness of GAs has nothing to do with the Turing-completeness of the {\it solutions} constructed by some GAs (such as some GAs dealing with Genetic Programming~\cite{koza89,tel94,np08}). It consists in showing the existence of a GA which can {\it simulate} any program (or technically, Turing machine) for any input it could receive. That is, when the initial configuration of that GA codifies that Turing machine and its input in some form, the final solution of the GA provides the output of that Turing machine for that input.

Achieving Turing-completeness requires representing all the internal memory of a program no matter how much it grows. Therefore chromosomes will be required to grow arbitrarily in the GA constructed for our proof, like they do in GA domains such as e.g. Genetic Programming when no size bounds are set. Given the way GAs behave exactly like programs in our proof, constraining the sizes of chromosomes would not make GAs easy to predict though, in the same way as predicting the output of programs using only a polynomial amount of memory, or executing only for an exponential number of steps, is decidable but PSPACE-hard or EXPTIME-hard, respectively (and thus, intractable if P$\neq$PSPACE and regardless any unproven result, respectively).

Note that if a computation model includes some component allowed to be {\it any program} then the model will very probably be Turing-complete. We could argue that GAs are general enough to let some components, such as its fitness function, be any program. If achieving the Turing-completeness required exploiting that total freedom to define some component in {\it any} way, then one could wonder if the Turing-completeness exists {\it only} for GAs with that high level of sophistication. Were that the case, many categories of simpler but widely used GAs could be non-Turing-complete and easily predictable, giving the Turing-completeness of GAs {\it in general} a marginal impact in practice. Quite on the contrary, our proof will show that GAs are Turing-complete even when all their components are defined in remarkably simple ways. In particular, the fitness function will just check for substring inclusion. This simplicity will show that, if we wished to define a particular category of non-Turing-complete and efficiently-predictable GAs by constraining the way GA components are defined, then constraints would be so strong as to nearly nullify the resulting category.

\subsection{Proving the Turing-completeness of genetic algorithms}\label{sec:proofgenetic}

In this section we prove the Turing-completeness of GA by constructing a GA capable of simulating any given Turing Machine for any given input, i.e. a GA being universal for Turing machines. 
Technically, we will present a GA that solves the problem of finding the complete computation {\it history} (i.e. sequence of all configurations iteratively reached during the computation) of any given Turing machine for any given input.
Before we begin with the description of this GA, we introduce some definitions and results that will help us build it.

Let $M = (Q,\Sigma,\Gamma,\delta,q_0,B,F)$ be a Turing Machine as defined in \cite[p. 327]{hopcroft}, that is: $Q$ is the set of states; $\Sigma$ is the input alphabet; $\Gamma$ is the tape alphabet; $\delta$ is the transition function; $q_0\in Q$ is the initial state; $B\in\Gamma$ is the blank symbol; and $F\subseteq Q$ is the set of final states (accepting states).
The Post Correspondence Problem (PCP) consists in, given a finite set $W$ of pairs $(a,b)$ of finite strings, find out if, for some finite sequence of pairs (where there may be several occurrences of the same pair), the string obtained by consecutively reading the first components of the pairs and the string read on the second ones coincide. The Modified Post Correspondence Problem (MPCP) adds the constraint that the first pair of the sequence is fixed.
%
In order to show the undecidability of MPCP, in \cite[pp. 401-412]{hopcroft} any pair $(M,w)$, where $M$ is any Turing Machine and  $w\in\Sigma^*$ is its input, is mapped into a set of MPCP pairs $T$ such that $M$ halts for $w$ iff the answer of MPCP for $T$ is yes. We will use this deterministically generated finite set $T$ of pairs in our construction next.



This set of pairs $T$ can only be generated if the TM never moves left from the initial position and never writes blanks. Luckily, for every TM there is an equivalent TM with these restrictions~\cite{hopcroft}.
Given the way $T$ is constructed, we will have a pair to represent the initial state and input of $(M,w)$, a closing pair to represent the end of the computation, pairs to represent the accepting states $q_f$ of $M$ and that will simulate the consumption of every symbol in the tape, a $(x,x)$ pair for each $x\in\Gamma$, a $(\#,\#)$ pair for the separator symbol $\#$ (these symbols separate the representation of each full TM configuration from the one reached in the next computation step), and a pair to represent every transition in $\delta$. Furthermore, if the first pair of a MPCP partial solution is the one that represents the initial state of the machine, then there is one and only one pair $t\in T$ able to extend that partial solution. This allows the extension of the partial solution to effectively emulate the computation of $(M,w)$.

In order to prove the undecidability of PCP, in~\cite{hopcroft} the halting problem for Turing machines is reduced to MPCP, and MPCP is reduced to PCP. Thus, if PCP were decidable, so would be the halting problem (which is undecidable). As mentioned before, this proof uses the pairs in $T$ to establish that MPCP has a solution if and only if $M$ stops with $w$ as its input. Moreover, the shortest possible solution for MPCP will represent the complete computation history of $M$ for input $w$.

With these ingredients, we can design a GA that tries to solve MPCP for input $T$ by sequentially finding the next pair (the next step of the computation) until MPCP is solved. We will prove that, for every computation step of the represented TM, our GA converges to the next step of the computation. This means that the GA can successfully emulate the behavior and tape status of every possible $(M,w)$ tuple.

The idea is that this GA will mimic the way genetic programming builds structured individuals (in this case just linearly structured, i.e. sequences with different sizes) in order to design an individual that arranges the pairs $t_i$ from $T$ into incremental partial solutions of MPCP until a complete solution is found. We will employ the operational semantics for GAs to describe this algorithm. Its population will consist of just one individual. This individual is represented as a sequence of elements we will call \textit{tiles}, and we should interpret $tile_i$ as the computational representation of $t_i \in T$ (i.e. the representation of $t_i$ in the GA domain).

The fitness function always returns the number of correctly arranged tiles and stores the best partial solution found so far (returning this solution is necessary only at the end of the computation). Both the selection and crossover stages are unnecessary and will not change anything in the individuals. The changes from the mutation stage only affect the end of the tile sequence of the individual, i.e. the end of the program execution history the GA is building: if the last tile is incorrect, then it is substituted by another random one; if it is correct, then a new random one is added afterwards, expanding the sequence.

With these elements we can guarantee, for every execution step, that the GA will always have convergence in probability to add the next correct execution step. However, by introducing a blacklist of tiles via \textit{Extra} to filter the ones already discarded in the current step of the MPCP, we can guarantee regular convergence. In order to accelerate the convergence of the implementation, the special pair of $T$ denoting the beginning of the sequence may have its corresponding tile removed from those that can be randomly added to the individual. Finally, the GA will only stop when the tile for the closing pair has been placed correctly.

Note that this is an extremely simple GA, in particular with a single individual population that only varies via mutation. This makes it akin to simple trajectory-based methods like Simulated Annealing (SA) or Iterated Local Search (ILS).


Now we define the data structures (not to be confused with the actual GA variables being instances of them; boldface and italics are used, respectively, to distinguish them) and the functions used in our operational semantics:

\subsubsection*{Data Structures}
\begin{itemize}
\item \textbf{Tile} is a pair of strings. It is used to represent pairs $t_i \in T$.
\item \textbf{Problem} is an array of elements of type \textbf{Tile} and an integer, which represent the set $T$ of pairs in the MPCP codification of the input $(M,w)$ and $|T|$ respectively.
\item \textbf{Solution} consists of a linked list of \textbf{Tile} elements, which represents an ordered arrangement of the pairs that code the instance of MPCP.
\item \textbf{F/P/W} is just one integer. It counts the number of correct tiles chained so far in the only element of \textit{Sols}.
\item \textbf{Extra} is an array of Boolean values. It represents a blacklist of tiles that resets every time a new correct one is added to the partial solution, and filters out the incorrect ones tried so far for the current iteration.
\end{itemize}

\subsubsection*{Function Definitions}
\begin{itemize}
\item $\C\llbracket A \rrbracket$ 
returns an instance of \textbf{Extra} with every value set to true.
\item $\PP\llbracket p \rrbracket_s$
processes $(M,w)$ to create $T$ as in \cite[pp. 401-412]{hopcroft}, then codes every pair in $T$ as instances of \textbf{Tile} and gives a \textbf{Problem} instance with the initial tile in the first position of the array, the closing tile in the second position and all the others in the next $|T|-2$ positions. The integer value of \textbf{Problem} is set to $|T|$.
\item $\GG\llbracket \rrbracket_s$
gives a linked list with 2 elements, the first one is the first element of the array \textit{Prob}\footnote{The representation of the problem to solve, in this case the list of all MPCP tiles.}, and the second one is chosen randomly among all other positions of \textit{Prob} with index lower than the integer stored. It also returns an instance of \textbf{Extra} equal to \textit{Extra} with the corresponding position set to false.
\item $\Select\llbracket \rrbracket_s$
always returns the only element in \textit{Sols} and \textit{Extra}.
\item $\Cross\llbracket \rrbracket_s$
always returns the only element in \textit{Sols} and \textit{Extra}.
\item $\Mutate\llbracket \rrbracket_s$ is defined as follows.
If \textit{FPW} equals the number of tiles in \textit{Best}, then it appends a new random tile from \textit{Prob} (with index $i\in[1,N-1]$, where $N$ is the stored integer, and the first index of the array is 0), and then returns the augmented list of tiles and an instance of \textbf{Extra} with every value set to true except for the value with index $i$ (this resets the blacklist).\\
Otherwise, a new tile from  \textit{Prob} is chosen randomly with index $i\in[1,N-1]$ where $Extra[i]$ is true. The returned values are a linked list of tiles consisting of a modified version of \textit{Best} with $Prob[i]$ at the end, and an instance of \textbf{Extra} equal to \textit{Extra} except for the value with index $i$, which is set to false.
\item $\EvaA\llbracket \rrbracket_s$
concatenates all the strings from the tiles of \textit{Sols} into two strings $a$ and $b$, where $a$ is the concatenation of the first elements of each tile in the order given by \textit{Sols} and $b$ is analogous for the second elements.\\
If \textit{FPW} has not been initialized but $b$ cannot be expressed as $a$ followed by other symbols, then the function returns 1 and \textit{Extra}.\\
If \textit{FPW} has not been initialized and $b$ can be expressed as $a$ followed by other symbols, then the function returns 2 and an instance of \textbf{Extra} with every element set to true.\\
Otherwise, the function returns \textit{FPW} and \textit{Extra} if $b$ cannot be expressed as $a$ followed by other characters, and returns $FPW+1$ and an instance of \textbf{Extra} with every element set to true if it can.
\item $\EvaB\llbracket \rrbracket_s$
returns the first \textit{FPW} tiles of \textit{Sols} linked in the same order.
\item $\SC \llbracket \rrbracket_s$
only stops if the number of tiles in \textit{Best} equals \textit{FPW} and the last tile of \textit{Best} equals $Prob[1]$ (the closing tile as defined in $\PP\llbracket p \rrbracket_s$).
\end{itemize}

\subsubsection*{Proof of Turing-completeness}

\begin{theorem}
 There is a genetic algorithm that, given any Turing Machine $M$ and any input $w$, is able to simulate every step of the computation of $M$ for $w$. Thus, genetic algorithms are Turing-complete.
\end{theorem}

\begin{proof}
To prove this result, we will consider the GA described above and use induction over the number of computed steps. Let $a$ be the concatenation of the first elements of each tile in $Best$ respecting their order, $b$ the concatenation of the second elements and let $a_i, b_i$ be the strings between the $(i-1)$-th and the $(i)$-th separators $\#$ in $a$ and $b$ respectively.

The induction hypothesis (IH) will be the following: if a computation of $(M,w)$ takes $n$ transitions then, at some point, $b$ will have $n+2$ separators $\#$, no more symbols after the $(n+2)$-th separator, and each $b_i \subset b$ will correctly represent the state, pointer, and tape of $M$ after $i$ computation steps.

\textbf{Induction Basis: } For $n = 0$ we have to prove that, at some point, $\exists\; b_0 \subset b$ such that $b_0$ represents the original configuration of $(M,w)$. This is granted after the first evaluation, since the first tile of the chromosome is determined to be the first pair of the MPCP by the algorithm.

\textbf{Induction Step: } For $n = k+1$ we use the (IH) and see how the algorithm behaves at that point in time. $b_k$ is a string with a finite set of symbols. The construction of the pair set $T$ seen in \cite[pp. 401-412]{hopcroft} implies that one and only one tile can augment a partial solution of MPCP. Also, there are no tiles whose first or second elements contain 0 symbols.

Thus, if the described GA always finds the next tile to augment its current partial solution stored in $Best$, then after adding $|b_k|+1$ tiles (where $|b_k|$ is the number of symbols present in string $b_k$) we will have added at least one more separator to $a$ (to match the one present in $b$), which implies adding it to $b$, since the only tiles that have a separator are $(\#,\#)$ and $(q_f\#\#,\#)$.

All that remains to be proved is that the added tiles  have correctly computed $b_k$, and that the described GA always finds the next tile to augment $Best$. The first property is given in~\cite[pp. 401-412]{hopcroft}, whereas the second one is proved in the following lemma.

\begin{lemma}

The GA described above always finds the next tile to augment its current partial solution stored in $Best$.

\end{lemma}

\begin{proof}

The construction of $T$ implies that, for every partial solution $c$ of the MPCP instance, there exists one and only one $t_i\in T$ such that $c$ followed by $t_i$ is also a partial solution \cite[pp. 401-412]{hopcroft}. By applying this property to $Best$ once it has been initialized, we get that there is one and only one $\textit{Prob}[i]\in \textit{Prob}$ such that the concatenation of $Best$ and $\textit{Prob}[i]$ is also a partial solution.

The mutation step chooses a random, not initial, and not blacklisted tile from \textit{Prob}. In the first iteration after modifying $Best$, the probability of choosing the correct tile is $\frac{1}{|Prob|-1}$. The divisor decreases by one unit for every blacklisted tile, and each erroneous choice blacklists one tile. In the worst case scenario, after $|Prob|-2$ iterations the probability of choosing the right tile is $1$.

\end{proof}

This completes the proof of the induction step. Thus, we have proved that the described GA can correctly emulate every step of the computation of any given $(M,w)$.

\end{proof}

Analogously, it can be proved that our GA halts iff $(M,w)$ does, which makes the Halting Problem undecidable for SLS computation in general; but this is much simpler to prove by directly applying the properties of $T$ as follows.
If $(M,w)$ halts, then a tile that represents an accepting state in its first string will be added to our partial solution. If this is a $(q_f\#\#,\#)$ tile, then $\SC \llbracket \rrbracket_s$ will return \textbf{tt} and our GA will halt. If it is not that kind of tile, then the GA will keep on adding tiles that eliminate the symbols of the tape like $(X q_f,q_f)$, $(X q_f Y,q_f)$, or $(q_f Y,q_f)$ until the only viable option is a $(q_f\#\#,\#)$ tile. A more detailed explanation of this elimination process 
can be seen in \cite[pp. 401-412]{hopcroft}.

It is worth noting that the proposed universal GA is particularly simple. Essentially, it lies in just two key ideas: (a)~when the mutation operator changes a chromosome (the only one in the population), it just modifies its ending part, which is replaced or expanded by some sequence randomly taken from a given finite set of possible sequences (the tiles); and (b)~the fitness function, as well as the condition to decide between replacing or expanding in (a), just checks if some sequence is a prefix of another one and for how long. The mechanics of this GA are particularly simple, and by no means they embed the internals of the behavior of a TM (tape, pointer moves, states, etc.). The TM to be simulated by the GA is actually received in the form of the finite set mentioned in~(a).
Note that using any Universal Turing Machine (UTM)\footnote{A Turing Machine that, given any TM and an input, simulates the computation of the latter TM for that input, see e.g. \cite[p. 20]{arora}.} requires receiving the TM to be simulated by it codified into some arbitrary notation, and similarly, simulating any TM with our GA requires receiving the TM codified into an appropriate notation, in particular that finite set mentioned in~(a). Although this set can be manually defined for each simulation case, there is actually an automatic way to obtain it from a TM defined with its most typical notation: converting it into a set of MPCP tiles as described in~\cite{hopcroft}.

Note that we could trivially make a deterministic universal GA from our GA by removing its randomness. We could do it just by making the GA try the different tiles to be added to the chromosome in any arbitrary deterministic order, instead of randomly (the blacklist could be trivially used to know which tile must be tried next). 




Finally, we show that a more memory-efficient (but less simple) implementation can be created by eliminating from \textit{Sols} those tiles whose information is no longer relevant to the evolution of the solution. If some symbols on the first parts of the tiles have already been completely matched in the concatenation of the second parts, then they no longer provide information to determine the next tile to be appended to \textit{Sols}. Moreover, if a symbol in the concatenation of the first parts has at least two separator symbols $\#$ to its right in the concatenation (not necessarily together or right next to it), then eliminating it still preserves the entirety of the TM configuration in the last computation step (recall that each full TM configuration is surrounded by a $\#$ symbol at each side). Therefore, we can safely eliminate from \textit{Sols} all the tiles whose symbols in the first parts have already been completely matched and have at least two \# symbols to their right in the concatenation of first parts.

This implementation would require careful modifications in \textit{Extra} and the functions previously defined, in particular to keep track of how many symbols from the concatenation of the second parts of \textit{Sols} need to be skipped by $\EvaA\llbracket \rrbracket_s$ (because they already matched symbols that have been removed from the concatenation of the first parts), as well as to take into account that we no longer keep the full computation history, but just the TM configuration at the last step and part of the configuration at the current one. Note that, by removing tiles this way, \textit{Sols} will never contain symbols from the representation of more than two TM configurations. The stop condition would now require that the last tile is $Prob[1]$ and that the concatenations of the first and second parts match if we skip the already-matched symbols of the concatenation of the second parts. The remaining functions would be trivially adapted.

By trimming unnecessary tiles this way, the representation size of \textit{Sols} (and thus, the representation size of our only individual in the GA) does not grow with the number of simulated TM steps as in our original construction, but it is just proportional to the maximum amount of memory used by the simulated TM during its execution. 
This independence between the size of the chromosome and the number of simulated TM steps is not required at all to achieve the Turing-completeness of GAs. However, it is needed for other Complexity-related properties, such as e.g. the property that GAs where chromosomes have polynomial size can simulate any TM running in polynomial space. Note that this would not be possible if the full history was stored in the chromosome, as polynomial-space TMs can take exponential time (i.e. have {\it exponential-size} histories). On the contrary, tile-trimming GAs with polynomial-size chromosomes can simulate all TMs running in polynomial space, so even very basic semantic properties of GAs with polynomial-size chromosomes are PSPACE-hard to decide. 




\section{Proving Turing-completeness for other instances of the general form}\label{sec:others}

The Turing-completeness proof presented in the previous section can be adapted to work with other instances of SLS computation. Of course, not all instances of SLS algorithms are Turing-complete; however, we claim that most other popular methods different from GA are too. Next we informally illustrate how we could prove it for the other two methods instantiated in this paper.

Let us consider Ant Colony Optimization. From a TM and input $(M,w)$ we can build the set $T$ of MPCP tiles as before. We can construct an ACO instantiation such that, if it is applied to that MPCP instance, then it necessarily tends to solve it, thus developing on the fly the execution of $M$ for $w$. Since Turing-completeness requires unlimited memory,
an infinite graph to be traversed by the ants is necessary in ACO, or more precisely, a finite graph (in particular, a tree) that can be iteratively extended during the ACO execution up to any needed finite size on demand, without limit. Each edge represents taking some tile of the MPCP instance, and each path from the root to any leaf represents a partial solution (sequence of tiles). In the beginning of the execution of ACO, this tree just consists of a single node. When an ant reaches a leaf, the ant finishes its journey and the path is reinforced via pheromones only if the tile sequence of the path consists only of legal moves (i.e. if one of the two strings read through the sequence of tiles is a prefix of the other, as we did for GA). In that case, right after removing the ant new edges are created from the leaf (thus no longer a leaf), one for each available tile, which will eventually let subsequent ants go one step further.


This way, ACO encourages ants to follow legal paths according to the tiles and extend them. Pheromone trails are reinforced only on the correctly computed tiles (edges). We assume that each newly created edge starts the execution with a 0 pheromone trail, pheromones do not disappear after any number of iterations,
and probabilities to select each edge are directly proportional to the amount of pheromones in all available edges. We can see that, in this setting, there is convergence in probability to eventually reach any step of the TM computation. By using the same stop criterion used for the Turing-complete GA, we again achieve Turing-completeness.

Proving the Turing-completeness of Particle Swarm Optimization is more closely related to the proof for GA than the one for ACO. In addition to the standard adaptations of PSO to deal with a discrete domain, we need the solution vectors to be arbitrarily enlarged on demand to represent arbitrarily long representations of the full TM memory state, as we did for GA.
Moreover, similarly as the mutation operator of GA was restricted to affect only the last part of the solution (i.e. the part it grew from), the two forces governing the movement of particles in PSO are defined to affect only specific scopes as follows: The force that makes particles move closer to the best particle will make all particles necessarily copy all the components (tiles) of the best solution; and the force that makes particles randomly move will make particles create random additions in a new component (tile) of its vector, used to let the sequences of MPCP tiles enlarge. This way, the particle that copies the former best solution (i.e. becomes it) and next randomly adds the suitable addition to its vector (which must be the legal tile) becomes the new best particle, the one to be followed next. Again, this process converges in probability to eventually creating the sequence of tiles representing the corresponding execution of the TM.

Similarly to the GA case, blacklists can be introduced in the previous ACO and PSO constructions to turn the convergence in probability into regular convergence. This way, the maximum number of steps up to termination would be bounded by a function over the number of steps actually needed by the TM to run for its input. 
Also, unnecessary elements of the constructed trees and solution vectors of ACO and PSO, respectively, could be trimmed as we did in the alternative GA construction described in the last paragraphs of the previous section.


\section{Conclusions and future work}\label{sec:conclusions}

We have presented a fully formal language which, by means of its common structure and its parametric constructions, lets us define any stochastic local search method to the best of our knowledge. In order to illustrate its flexibility, its instances for three algorithm families with very different mechanics (GA, PSO, ACO) have been presented.By using our common and standardized language in the definition of new and existing algorithms, the functional relationships between algorithmic choices and problem features could be more easily investigated. Results from different researchers, working with different algorithms, could be seamlessly classified and added into a common knowledge pool.

We have also formally proved the Turing-completeness of the GA instance, which implies the Turing-completeness of the general model it instantiates, and thus the Turing-completeness of stochastic local search methods in general. Besides, we have sketched how this Turing-completeness proof can be easily adapted to the other two instances presented, which supports the idea that many other important families will be Turing-complete too. 

By Rice's theorem, the Turing-completeness of a method implies the undecidability of all its non-trivial semantic properties (i.e. the ones concerning the relations it computes between its inputs and its outputs, and being fulfilled neither always nor never). This undecidability means that, for a given property, no algorithm exists that decides whether the property is satisfied or not. This result constitutes an important limitation to any attempt to predict the behavior of these methods. For example, it implies the undecidability, in general, of whether GA, ACO, or PSO instances, allegedly designed to solve some optimization problem, will eventually produce solutions of some desired kind (e.g. optimal solutions, solutions always reaching some performance ratio, solutions reaching some performance ratio on average, solutions fulfilling some given structural condition, etc.). These semantic properties may still be proved for particular cases (e.g. a given SLS method solving a particular problem or problem instance), as well as for subsets of cases disabling Turing-completeness by constraining the general freedom of the method or the problem. However, as we pointed out in the paper, removing the Turing-completeness (e.g. by not letting GA chromosomes grow unboundedly) is not enough,by itself, to enable {\it efficient} property detection in general, as it may only turn the undecidability into intractability (e.g. PSPACE-hardness).

Regarding our lines of future work, we wish to use our model to develop a fully formal taxonomy of stochastic local search  methods. Different levels of that taxonomy would be reached by instantiating the General Form up to a higher or lower level of abstraction. Additionally, we wish to investigate the complexity of semantic property identification in general when chromosomes cannot grow {\it at all}. 
%
%
We suspect that we would still have PSPACE-hardness in this case, and note that this intractability would be consistent with known convergence results in the SLS field. For instance, SA is guaranteed to reach the optimum solution if we can proceed at a slow enough pace, in particular taking exponential time in some bad cases ~\cite{nolte1996simulated}. This makes the heuristic search take, in practical terms, as much time as an exhaustive search over the corresponding (exponential-size) search space.

Finally, we wish to systematically investigate what restrictions in the definition of the {\it elements} of a heuristic, beyond the representation size of algorithm entities, make the difference between enabling Turing-completeness for that method and not enabling it, with special attention to non population-based heuristics due to their simplicity.




\section*{References}

\bibliography{tse_bib}

@inproceedings{rrrUC07,
  title={Using river formation dynamics to design heuristic algorithms},
  author={Rabanal, Pablo and Rodr{\'\i}guez, Ismael and Rubio, Fernando},
  booktitle={International Conference on Unconventional Computation},
  pages={163--177},
  year={2007},
  organization={Springer}
}

@book{ga06,
  title={Genetic algorithms},
  author={Goldberg, David E.},
  year={2006},
  publisher={Pearson Education}
}

@incollection{pso11,
  title={Particle swarm optimization},
  author={Kennedy, James},
  booktitle={Encyclopedia of machine learning},
  pages={760--766},
  year={2011},
  publisher={Springer}
}

@inproceedings{aco99,
  title={Ant colony optimization: a new meta-heuristic},
  author={Dorigo, Marco and Di Caro, Gianni},
  booktitle={Proceedings of the 1999 Congress on Evolutionary Computation, CEC'99},
  volume={2},
  pages={1470--1477},
  year={1999},
  organization={IEEE}
}

@article{aco06,
  title={Ant colony optimization},
  author={Dorigo, Marco and Birattari, Mauro and Stutzle, Thomas},
  journal={IEEE computational intelligence magazine},
  volume={1},
  number={4},
  pages={28--39},
  year={2006},
  publisher={IEEE}
}

@book{evol97,
 editor = {Back, Thomas and Fogel, David B. and Michalewicz, Zbigniew},
 title = {Handbook of Evolutionary Computation},
 year = {1997},
 isbn = {0750303921},
 publisher = {IOP Publishing},
 address = {Bristol, UK},
}

@book{swarm01,
  title={Swarm intelligence},
  author={Kennedy, James and Eberhart, Russell C. and Shi, Yuhui},
  year={2001},
  publisher={Morgan Kaufmann}
}

@article{DCG05,
 author={Ducatelle, Frederick and Di Caro, Gianni and Gambardella, Luca Maria},
 title     = {Using Ant Agents to Combine Reactive and Proactive Strategies for
              Routing in Mobile Ad-hoc Networks},
 journal   = {International Journal of Computational Intelligence and Applications},
 volume    = {5},
 number    = {2},
 pages     = {169--184},
 year      = {2005},
 issn	   = {1469-0268},
 aadoi       = {10.1142/S1469026805001556},
 publisher = {World Scientific Publishing Co}
}

@article{KaA11,
 author = {D. Karaboga and B. Akay},
 title = {A Modified Artificial Bee Colony ({ABC}) Algorithm for Constrained Optimization Problems},
 journal = {Applied Soft Computing},
 volume = {11},
 number = {3},
 year = {2011},
 issn = {1568-4946},
 pages = {3021--3031},
 aadoi = {10.1016/j.asoc.2010.12.001},
 publisher = {Elsevier}
}

@article{QWZ13,
 author = {C.Y. Qiu and C.L. Wang and X.Q. Zuo},
 title = {A novel multi-objective particle swarm optimization with K-means based global best selection strategy},
 journal = {International Journal of Computational Intelligence Systems},
 volume = {6},
 number = {5},
 year = {2013},
 issn = {1875-6883},
 pages = {822--835},
 aadoi = {10.1080/18756891.2013.805584},
 publisher = {Taylor \& Francis}
}

@article{kirkpatrick1983osa,
  title={{Optimization by Simulated Annealing}},
  author={S. Kirkpatrick and C.D. Gelatt Jr. and M.P. Vecchi},
  journal={Science},
  volume={220},
  number={4598},
  pages={671},
  year={1983}
}

@ARTICLE{pas09,
   author ="V.T. Paschos",
   title ="An overview on polynomial approximation of {NP}-hard problems",
   journal ="Yugoslav Journal of Operations Research",
   year ="2009",
   volume = "19",
   number = "1",
   pages ="3--40"
}

@book{cut80,
  title={Computability: An Introduction to Recursive Function Theory},
  author={N. Cutland},
  year={1980},
  publisher={Cambridge University Press}
}

@article{rice,
author = "H. G. Rice",
title = "Classes of recursively enumerable sets and their decision problems",
journal = "Transactions of the American Mathematical Society",
volume = "74",
pages = "358-366",
year = "1953",
}

@inproceedings{overfitting,
author = {P. Rabanal and I. Rodr{\'\i}guez and F. Rubio},
title = "Assessing Metaheuristics by Means of Random Benchmarks",
booktitle = "International Conference on Computational Science 2016 (ICCS 2016),  Procedia Computer Science, vol.80",
pages = {289-300},
year = {2016}
}

@article{sor15,
author = "K. S{\"{o}}rensen",
title = "Metaheuristics---the metaphor exposed",
journal = "International Transactions in Operational Research",
volume = "22",
number = "1",
pages = "3--18",
year = "2015",
}

@BOOK{dejong,
  AUTHOR        = {{De Jong}, Kenneth Alan},
  BOOKTITLE     = {Optimization},
  ISBN          = {0262041944},
  PAGES         = {250},
  PUBLISHER     = {MIT press},
  TITLE         = {{Evolutionary Computation: A Unified Approach}},
  YEAR          = {2006},
}

@article{kang,
  title={Unification and diversity of computation models for generalized swarm intelligence},
  author={Kang, Qi and An, Jing and Wang, Lei and Wu, Qidi},
  journal={International Journal on Artificial Intelligence Tools},
  volume={21},
  number={03},
  pages={1240012},
  year={2012},
  publisher={World Scientific}
}

@article{f5,
  title={Distributed evolutionary algorithms and their models: A survey of the state-of-the-art},
  author={Gong, Yue-Jiao and Chen, Wei-Neng and Zhan, Zhi-Hui and Zhang, Jun and Li, Yun and Zhang, Qingfu and Li, Jing-Jing},
  journal={Applied Soft Computing},
  volume={34},
  pages={286--300},
  year={2015},
  publisher={Elsevier}
}

@article{f6,
  title={Swarm-based metaheuristics in automatic programming: a survey},
  author={Olmo, Juan L and Romero, Jos� R and Ventura, Sebasti�n},
  journal={Wiley Interdisciplinary Reviews: Data Mining and Knowledge Discovery},
  volume={4},
  number={6},
  pages={445--469},
  year={2014},
  publisher={Wiley Online Library}
}

@InProceedings{Koza89,
  author =       "J. R. Koza",
  title =        "Hierarchical genetic algorithms operating on
                 populations of computer programs",
  editor =       "N. S. Sridharan",
  volume =       "1",
  pages =        "768--774",
  booktitle =    "Proceedings of the Eleventh International Joint
                 Conference on Artificial Intelligence IJCAI-89",
  year =         "1989",
  address =      "Detroit, MI, USA",
  publisher =    "Morgan Kaufmann",
  publisher_address = "San Mateo, CA, USA",
  month =        "20-25 " # aug,
  acmid =        "1623877",
}

@inbook{arora,
    author    = "Sanjeev Arora and Boaz Barak",
    title     = "Computational Complexity, A Modern Approach",
    year      = "2009",
    publisher = "Cambridge University Press"
}

@conference{aco,
  author       = {A. Colorni and M. Dorigo and V. Maniezzo},
  title    = {Distributed Optimization by Ant Colonies},
  year         = 1991,
  pages        = "134--142",
  address      = {Paris, France},
  organization = {Actes de la Première Conférence Européenne sur la Vie Artificielle},
  publisher    = {Elsevier Publishing}
}

@inbook{hopcroft,
    author    = "John E. Hopcroft and Ranjeev Motwani and Jeffrey D. Ullman",
    title     = "Introduction to Automata Theory, Languages and Computation",
    year      = "2001",
    edition   = "3rd",
    publisher = "Pearson Education"
}

@inbook{nielson,
    author    = "Hanne Riis Nielson and Flemming Nielson",
    title     = "Semantics With Applications: A Formal Introduction",
    year      = "1999",
    publisher = "Wiley Professional Computing"
}

@inproceedings{tel94,
author="A. Teller",
title="Turing completeness in the language of genetic programming with indexed memory",
booktitle="Proceedings of the First IEEE Conference on Evolutionary Computation. IEEE World Congress on Computational Intelligence",
pages="136--141",
PUBLISHER = "IEEE Press",
year="1994"}

@inproceedings{np08,
author="A. Naidoo and N. Pillay",
title="Using genetic programming for {T}uring machine induction",
booktitle="EuroGP'08 Proceedings of the 11th European conference on Genetic programming",
pages="350--361",
PUBLISHER = "Springer-Verlag",
year="2008"}

@article{pnd09,
 author = {P.C. Pinto and A. N\"{a}gele and M. Dejori and T.A. Runkler and J.M.C. Sousa},
 title = {Using a Local Discovery Ant Algorithm for Bayesian Network Structure Learning},
 journal = {Trans. Evol. Comp},
 issue_date = {August 2009},
 volume = {13},
 number = {4},
 month = aug,
 year = {2009},
 pages = {767--779},
publisher = {IEEE Press},
}

@article{rrr17,
 author = {I. Rodr{\'i}guez and P. Rabanal and F. Rubio},
 title = {How to make a best-seller: Optimal product design problems},
 journal = {Applied Soft Computing},
 volume = {55},
 year = {2017},
 pages = {178--196},
 publisher = {Elsevier}
}

@inproceedings{cq07,
  author    = {C. Lin and Q. Feng},
  title     = {The Standard Particle Swarm Optimization Algorithm Convergence Analysis and Parameter Selection},
  booktitle = {Third International Conference on Natural Computation, {ICNC} 2007, Volume 3},
  pages     = {823--826},
  year      = {2007}
}

@article{jly07,
  author    = {M. Jiang and Y. Luo and S. Yang},
  title     = {Stochastic convergence analysis and parameter selection of the standard particle swarm optimization algorithm},
  journal   = {Inf. Process. Lett.},
  volume    = {102},
  number    = {1},
  pages     = {8--16},
  year      = {2007},
}

@article{tre03,
  author    = {I.C. Trelea},
  title     = {The particle swarm optimization algorithm: convergence analysis and parameter selection},
  journal   = {Inf. Process. Lett.},
  volume    = {85},
  number    = {6},
  pages     = {317--325},
  year      = {2003},
}

@article{lwy16,
  title={Topology selection for particle swarm optimization},
  author={Q. Liu and W. Weiand H. Yuan and Z.H. Zhan and Y. Li},
  journal={Information Sciences},
  volume={363},
  pages={154--173},
  year={2016},
  publisher={Elsevier}
}

@inproceedings{ska15,
  author    = {M. Sekara and M. Kowalski and A. Byrski and B. Indurkhya and  M. Kisiel{-}Dorohinicki and D. Samson and T. Lenaerts},
  title     = {Multi-pheromone ant Colony Optimization for Socio-cognitive Simulation Purposes},
  booktitle = {Proceedings of the International Conference on Computational Science, {ICCS} 2015},
  pages     = {954--963},
  year      = {2015}
}

@inproceedings{can99,
author = {E. Cant\'{u}-Paz},
title = {Topologies, Migration Rates, and Multi-Population Parallel Genetic Algorithms},
year = {1999},
publisher = {Morgan Kaufmann Publishers Inc.},
booktitle = {{GECCO}'99, Proceedings of the 1st Annual Conference on Genetic and Evolutionary Computation - Volume 1},
pages = {91--98}
}

@article{dmq11,
  author    = {S. Das and S. Maity and B.Y. Qu and P.N. Suganthan},
  title     = {Real-parameter evolutionary multimodal optimization - {A} survey of the state-of-the-art},
  journal   = {Swarm and Evolutionary Computation},
  volume    = {1},
  number    = {2},
  pages     = {71--88},
  year      = {2011},
}

@article{jws13,
  author    = {L. Jiao and H. Wang and R. Shang and F. Liu},
  title     = {A co-evolutionary multi-objective optimization algorithm based on direction vectors},
  journal   = {Inf. Sci.},
  volume    = {228},
  pages     = {90--112},
  year      = {2013},
}

@article{zl07,
  author    = {Q. Zhang and H. Li},
  title     = {{MOEA/D:} {A} Multiobjective Evolutionary Algorithm Based on Decomposition},
  journal   = {{IEEE} Trans. Evolutionary Computation},
  volume    = {11},
  number    = {6},
  pages     = {712--731},
  year      = {2007},
}

@Article{ma04,
author="R.T. Marler and J.S. Arora",
title="Survey of multi-objective optimization methods for engineering",
journal="Structural and Multidisciplinary Optimization",
year="2004",
month="Apr",
day="01",
volume="26",
number="6",
pages="369--395",
}

@inproceedings{kb07,
  author    = {D. Karaboga and B. Basturk},
title     = {Artificial Bee Colony {(ABC)} Optimization Algorithm for Solving Constrained Optimization Problems},
  booktitle = {Foundations of Fuzzy Logic and Soft Computing, 12th International Fuzzy Systems Association World Congress, {IFSA} 2007},
  series    = {Lecture Notes in Computer Science},
  volume    = {4529},
  pages     = {789--798},
  publisher = {Springer},
  year      = {2007},
}

@misc{molina2020comprehensive,
    title={Comprehensive Taxonomies of Nature- and Bio-inspired Optimization: Inspiration versus Algorithmic Behavior, Critical Analysis and Recommendations},
    author={Daniel Molina and Javier Poyatos and Javier Del Ser and Salvador Garc\'ia and Amir Hussain and Francisco Herrera},
    year={2020},
    eprint={2002.08136},
    archivePrefix={arXiv},
    primaryClass={cs.AI}
}

@Inbook{Hoos2015,
author="Hoos, Holger H.
and St{\"u}tzle, Thomas",
editor="Kacprzyk, Janusz
and Pedrycz, Witold",
title="Stochastic Local Search Algorithms: An Overview",
bookTitle="Springer Handbook of Computational Intelligence",
year="2015",
publisher="Springer Berlin Heidelberg",
address="Berlin, Heidelberg",
pages="1085--1105",
isbn="978-3-662-43505-2",
doi="10.1007/978-3-662-43505-2_54",
url="https://doi.org/10.1007/978-3-662-43505-2_54"
}

@article{hutter2009paramils,
  title={ParamILS: an automatic algorithm configuration framework},
  author={Hutter, Frank and Hoos, Holger H and Leyton-Brown, Kevin and St{\"u}tzle, Thomas},
  journal={Journal of Artificial Intelligence Research},
  volume={36},
  pages={267--306},
  year={2009}
}

@inproceedings{hutter2011sequential,
  title={Sequential model-based optimization for general algorithm configuration},
  author={Hutter, Frank and Hoos, Holger H and Leyton-Brown, Kevin},
  booktitle={International conference on learning and intelligent optimization},
  pages={507--523},
  year={2011},
  organization={Springer}
}

@article{lopez2016irace,
  title={The irace package: Iterated racing for automatic algorithm configuration},
  author={L{\'o}pez-Ib{\'a}nez, Manuel and Dubois-Lacoste, J{\'e}r{\'e}mie and C{\'a}ceres, Leslie P{\'e}rez and Birattari, Mauro and St{\"u}tzle, Thomas},
  journal={Operations Research Perspectives},
  volume={3},
  pages={43--58},
  year={2016},
  publisher={Elsevier}
}

@book{hoos2004stochastic,
  title={Stochastic local search: Foundations and applications},
  author={Hoos, Holger H and St{\"u}tzle, Thomas},
  year={2004},
  publisher={Elsevier},
  chapter={3},
  pages = {105 -- 136}
}

@article{rudolph1998finite,
  title={Finite Markov chain results in evolutionary computation: A tour d'horizon},
  author={Rudolph, G{\"u}nter},
  journal={Fundamenta informaticae},
  volume={35},
  number={1-4},
  pages={67--89},
  year={1998},
  publisher={IOS Press}
}

@article{mitra1986convergence,
  title={Convergence and finite-time behavior of simulated annealing},
  author={Mitra, Debasis and Romeo, Fabio and Sangiovanni-Vincentelli, Alberto},
  journal={Advances in applied probability},
  volume={18},
  number={3},
  pages={747--771},
  year={1986},
  publisher={Cambridge University Press}
}

@article{johnson2002convergence,
  title={On the convergence of generalized hill climbing algorithms},
  author={Johnson, Alan W. and Jacobson, Sheldon H.},
  journal={Discrete applied mathematics},
  volume={119},
  number={1-2},
  pages={37--57},
  year={2002},
  publisher={Elsevier}
}

@article{laumanns2004running,
  title={Running time analysis of multiobjective evolutionary algorithms on pseudo-boolean functions},
  author={Laumanns, Marco and Thiele, Lothar and Zitzler, Eckart},
  journal={IEEE Transactions on Evolutionary Computation},
  volume={8},
  number={2},
  pages={170--182},
  year={2004},
  publisher={IEEE}
}

@article{friedrich2015maximizing,
  title={Maximizing submodular functions under matroid constraints by evolutionary algorithms},
  author={Friedrich, Tobias and Neumann, Frank},
  journal={Evolutionary computation},
  volume={23},
  number={4},
  pages={543--558},
  year={2015},
  publisher={MIT Press}
}

@article{qian2019maximizing,
  title={Maximizing submodular or monotone approximately submodular functions by multi-objective evolutionary algorithms},
  author={Qian, Chao and Yu, Yang and Tang, Ke and Yao, Xin and Zhou, Zhi-Hua},
  journal={Artificial Intelligence},
  volume={275},
  pages={279--294},
  year={2019},
  publisher={Elsevier}
}

@inproceedings{nolte1996simulated,
  title={Simulated annealing and its problems to color graphs},
  author={Nolte, Andreas and Schrader, Rainer},
  booktitle={European Symposium on Algorithms},
  pages={138--151},
  year={1996},
  organization={Springer}
}



%

\end{document}